\documentclass{article} 
\usepackage{iclr2024_conference,times}

\usepackage{amsmath,amsfonts,bm}

\def\eqref#1{equation~\ref{#1}}

\def\floor#1{\lfloor #1 \rfloor}
\def\1{\bm{1}}

\def\eps{{\epsilon}}

\def\rvx{{\mathbf{x}}}

\def\mW{{\bm{W}}}

\DeclareMathAlphabet{\mathsfit}{\encodingdefault}{\sfdefault}{m}{sl}
\SetMathAlphabet{\mathsfit}{bold}{\encodingdefault}{\sfdefault}{bx}{n}

\newcommand{\R}{\mathbb{R}}

\usepackage{hyperref}
\usepackage{url}
\usepackage{subcaption}
\usepackage{multirow}
\usepackage{booktabs}
\usepackage{array}
\usepackage{graphicx}
\usepackage{tabulary}
\usepackage{wrapfig}
\hypersetup{
    colorlinks=true,
    linkcolor=blue,
    filecolor=magenta,      
    urlcolor=cyan,
    citecolor=blue,
}

\title{Navigating Scaling Laws: \\ Compute Optimality in Adaptive Model Training}
\iclrfinalcopy

\author{Sotiris Anagnostidis\thanks{Equal contribution. Correspondence to \texttt{\{sanagnos,gregorb\}@ethz.ch.}} \,, Gregor Bachmann$^{*}$, Imanol Schlag, Thomas Hofmann \\
Department of Computer Science\\
ETH Zürich
}

\newcommand{\Imagenet}[1]{\textit{ImageNet-{#1}k}}

\newcommand{\flexivit}{\textit{FlexiViT}~}

\begin{document}

\maketitle

\begin{abstract}
In recent years, the state-of-the-art in deep learning has been dominated by very large models that have been pre-trained on vast amounts of data. The paradigm is very simple: investing more computational resources (optimally) leads to better performance, and even predictably so; neural scaling laws have been derived that accurately forecast the performance of a network for a desired level of compute. This leads to the notion of a `compute-optimal' model, i.e. a model that allocates a given level of compute during training optimally to maximize performance. In this work, we extend the concept of optimality by allowing for an `adaptive' model, i.e. a  model that can change its shape during training. By doing so, we can design adaptive models that optimally traverse between the underlying scaling laws and outpace their `static' counterparts, leading to a significant reduction in the required compute to reach a given target performance. We show that our approach generalizes across modalities and different shape parameters. 
\end{abstract}

\section{Introduction}

Deep learning has gradually undergone a paradigm shift, where instead of training specialized models for a given task, a so-called frontier model is prompted/few-shoted/fine-tuned for different desired downstream tasks.
Frontier models are typically defined by their large-scale architectures, often rooted in the Transformer architecture~\citep{vaswani2017attention}, and their exposure to extensive and diverse data during their pre-training process, yielding remarkable advancements in both natural language understanding~\citep{openai2023gpt, kopf2023openassistant} and computer vision tasks~\citep{dehghani2023scaling, chen2023pali}. An inherent and pivotal feature of such models lies in their scalability, whereby their performance can be reliably predicted as a power law across the number of parameters and the volume of data or computational resources utilized~\citep{cortes1993learning, hestness2017deep, rosenfeld2019constructive, kaplan2020scaling}. These principles are succinctly encapsulated by the \emph{neural scaling laws} that motivate the choice of a particular model and dataset size given a fixed budget of training compute~\citep{hoffmann2022training}. The ability to accurately predict performance offers an undeniable reassurance in the often uncertain world of deep learning. It nevertheless, introduces an intimidating realization; 
\begin{center}
    \textit{Given a training scheme, a fixed further improvement in performance requires exponentially more compute or parameters.}
\end{center}
Finding solutions to address this issue becomes increasingly paramount, as staying competitive in the realm of deep learning increasingly depends on the availability of substantial computational resources. 
Delving deeper into the preceding statement, we highlight a pivotal assumption: \emph{the shape of the model}, and therefore the number of FLOPs for a forward pass, remains \emph{fixed throughout the training process}. By `shape' we refer to any characteristic of a model that can be smoothly varied throughout training without leading to strong deterioration in performance. Such a static approach (i.e. where every model shape remains fixed) may however not always be optimal. For example, it has already been observed that the optimal model size grows smoothly with the loss target and the compute budget~\citep{kaplan2020scaling}.

This paper challenges the assumption of a static model outlined above and explores adaptable training methodologies designed to \emph{surpass conventional scaling laws}. In other words, our aim is to achieve equivalent performance for a specified model with fewer computational resources (FLOPs) than initially projected. To that end, we adapt the shape of the model throughout training, allowing the optimal traversal between different scaling laws. This enables us to leverage the optimality of all shape configurations in different regions of compute, leading to a more efficient scaling of the model.

We train Vision Transformers~\citep{dosovitskiy2020image} and Language Models~\citep{radford2019language} and showcase how an adaptive training scheme can lead to substantial training FLOPs reduction, in some cases more than $50 \%$.
Our contributions can be summarized as follows:
\begin{itemize}
    \item We introduce a simple and effective strategy to choose when to adapt a model and traverse scaling laws, opting for the one that leads to the faster descent, i.e. maximum performance gain for the same amount of compute.
    \item We showcase the efficiency of our approach by optimally scheduling the patch size for ViTs as well as the context size for language models, leading to significant reductions in the required amount of compute to reach optimal performance.
    \item We further confirm the validity of our approach by adapting other shape parameters such as width, batch size, and overall training objective of a Vision Transformer.
\end{itemize}

\section{Related Work}

\begin{figure}
  \begin{subfigure}{0.76\linewidth}
    \centering
    \includegraphics[width=1\linewidth]{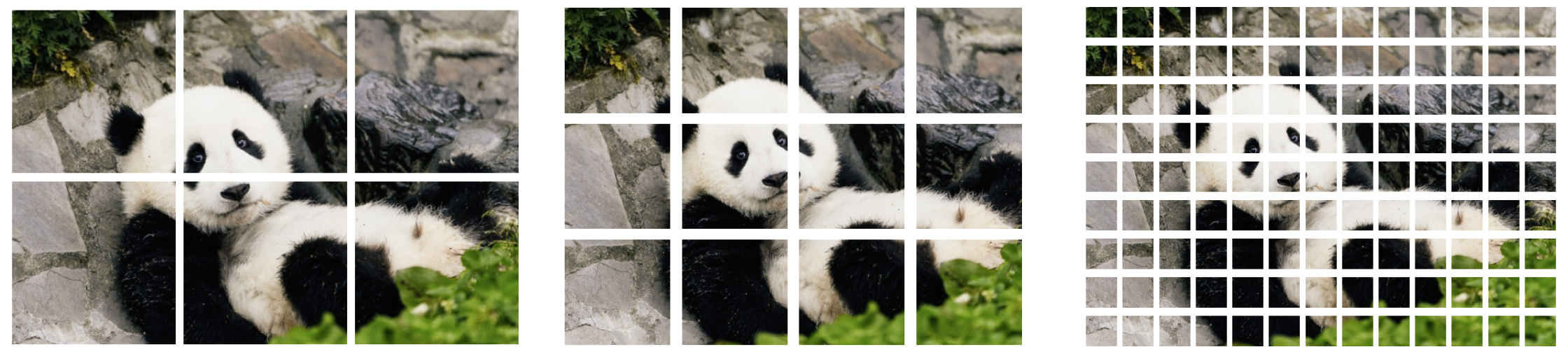}
    \caption{Patch sizes affect \textit{how} ViTs process input images.}
    \label{fig:patched-zaptos}
  \end{subfigure}%
  \hspace{1.5mm}
  \begin{subfigure}{0.215\linewidth}
    \centering
    \vspace{4mm}
    \includegraphics[width=1\linewidth]{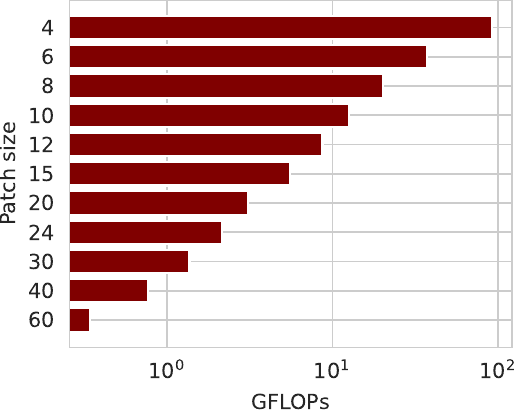}
    \caption{\textit{ViT-B} FLOPs.}
    \label{fig:flops_ps}
  \end{subfigure}%
  \caption{Patch sizes define (left) how images are processed, while (right) impacting the compute of a forward pass.}
  \vspace{-2mm}
\end{figure}

Neural scaling laws~\citep{cortes1993learning}, describe how a neural network's performance varies as a power law $E = a (P+d)^b + c$~where $P$ can be either the number of parameters in the model, the number of training samples or simply the number of FLOPs used for training~\citep{rosenfeld2019constructive}. Recently, scaling laws have been successfully demonstrated in a range of different applications, including language~\citep{kaplan2020scaling, hoffmann2022training} and vision~\citep{zhai2022scaling, bachmann2023scaling}, as well as numerous learning settings, including supervised training, generative modeling~\citep{henighan2020scaling} and transfer learning~\citep{hernandez2021scaling}. The predictive power of scaling laws has also been leveraged to determine compute-optimal models before training; the size of the \textit{Chinchilla} model and the number of training tokens were chosen based on the underlying scaling law and indeed, \textit{Chinchilla} outperformed its larger but sub-optimally trained counterpart \textit{Gopher}~\citep{hoffmann2022training}. The training of state-of-the-art models has also been guided by scaling laws built from training runs of smaller models \citep{openai2023gpt, team2023gemini}.

In this paper, we focus on the Transformer architecture and evaluate our methodology on both vision and natural language tasks.
While the Transformer has been the default model in language for years, ViTs have more recently established themselves as the predominant vision architecture for large-scale pretraining tasks~\citep{dehghani2023scaling}. Different from convolutions, a ViT initially partitions the input image into patches and processes these through self-attention and MLP blocks. This lack of inductive bias~\citep{smith2023convnets} can be partially overcome through the introduction of `soft' inductive bias, which proves to be beneficial, especially during the early phase of their training~\citep{d2021convit}. Similarly to their counterparts in natural language processing, ViTs also exhibit predictable scaling behavior~\citep{zhai2022scaling, dehghani2023scaling, alabdulmohsin2023getting}. 

In this work, we delve into models that feature adaptive `shape' parameters, specifically focusing on the patch size for image processing and model width. 
The concept of training with different patch sizes, which \citet{beyer2023flexivit} have explored, leads to a model robust to various patch sizes. 
Another common approach involves pre-training a ViT model at a lower resolution, followed by fine-tuning at a higher resolution while maintaining the same patch size \citep{dosovitskiy2020image, zhai2022scaling, alabdulmohsin2023getting}. 
Analogously, large language models (LLMs) can be pre-trained with a shorter and fixed context length and subsequently fine-tuned on longer ones~\citep{chen2023longlora, chen2023extending, tworkowski2023focused}.

Another way to change the model is by adding new parameters. Expanding a model under composable function-preserving operations has been a case of study for a long time in machine learning~\citep{ash1989dynamic, mitchell2023self}. The principal objective in this case is to accelerate training~\citep{kaddour2023no, geiping2023cramming}. Such expansion operations have also been proposed for the Transformer architecture~\citep{gesmundo2023composable, chen2022auto} and have exhibited notable training speed-ups~\cite{gong2019efficient, yao20232x, wang2023learning, lee2022towards, shen2022staged, li2022automated}. Apart from determining \emph{how} and \emph{where} in the model this expansion should occur, a primary challenge is to resolve \emph{when} to add new neurons. We advocate that an effective strategy for adjustments to the model shape should be informed by considerations of scaling laws and the performance gains achieved per additional unit of computational resources.

Orthogonal to our approach, various techniques have been proposed to accelerate both inference and training, particularly in the context of Transformer models. These methods encompass a spectrum of strategies, including weight quantization~\citep{dettmers2022llm, frantar2022gptq} and pruning weights and context~\citep{frantar2023sparsegpt, anagnostidis2023dynamic} among others. Specifically for ViTs,~\citet{bolya2022token} propose to merge tokens at different layers in the architecture and~\citet{dehghani2023patch} propose to pack sequences of tokens together to optimize hardware utilization. Additionally,~\citet{d2021convit} proposes to initialize ViTs differently, making them look more like convolutions. Other methods have also been proposed to beat scaling laws, including data pruning~\citep{sorscher2022beyond} or shaping models (depth vs width) more optimally~\citep{alabdulmohsin2023getting}. These approaches are supplementary to our methodology and can be effectively employed in conjunction to further enhance the efficiency of the training process.

\section{ViTs and Optimal Patch Sizes}
\label{sec:patch-size}

We first focus the discussion on Vision Transformers --- the de-facto dominant architecture for vision --- and the choice of patch size to introduce the notion of an adaptive model. In Sec.~\ref{sec:other_domains} we will showcase how our strategy can also be leveraged to train language models with adaptive context lengths. ViTs process images $\rvx \in \R^{h \times w \times c}$ where $h$ and $w$ are the height and width of the image in pixels and $c$ is the number of channels. Images are `patchified' into a sequence of $n$ tokens based on a specified patch size $p \in \mathbb{N}$, where $n = \floor{w / p} \times \floor{h / p}$, leading to a representation $\rvx_{\text{patched}} \in \R^{n \times p^2c}$. We illustrate the effect of different patch sizes in Fig.~\ref{fig:patched-zaptos}. 
Each token is linearly embedded with learnable parameters $\bm{W}_{emb} \in \mathbb{R}^{p^2c \times d}$ where $d \in \mathbb{N}$ is the embedding dimension or width of the ViT. These embeddings are further enhanced with learnable positional encodings $\bm{W}_{pos} \in \mathbb{R}^{n \times d}$ which enable a ViT to learn the spatial structure of the tokens. The resulting embeddings are then processed by $L$ transformer blocks, consisting of a self-attention layer followed by an MLP that is shared across tokens. This specific structure of the architecture allows a ViT to generate predictions for token sequences of variable lengths, as is the case when dealing with images of different patch sizes. 

\begin{figure*}[!t]
  \begin{subfigure}{0.38\linewidth}
    \centering
    \includegraphics[width=\linewidth]{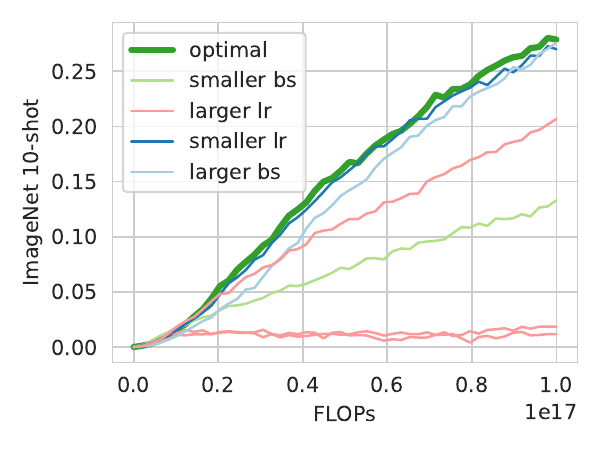}
    \caption{We optimize batch size, learning rate, and weight decay for each model configuration by running a greed search, for a small compute budget. More details are presented in the Appendix.}
  \end{subfigure}%
  \hspace{6.5mm}
  \begin{subfigure}{0.56\linewidth}
    \centering
    {\normalsize
    \begin{tabulary}{1.0\linewidth}{LCCCCRR}
        \toprule
        \multirow{3}{=}[3pt]{\centering \rotatebox{90}{\bf{Name}}} &
        \multirow{3}{=}[3pt]{\centering \rotatebox{90}{\bf{Width}}} &
        \multirow{3}{=}[3pt]{\centering \rotatebox{90}{\bf{Depth}}} &
        \multirow{3}{=}[3pt]{\centering \rotatebox{90}{\bf{Heads}}} &
        \multirow{3}{=}[3pt]{\centering \rotatebox{90}{\shortstack[c]{\bf{\begin{tabular}{@{}c@{}}Param \\ (M)\end{tabular}}}}} &
        \multicolumn{2}{c}{\bf{GFLOPs}} \\
        \cmidrule[0.5pt]{6-7}
         &  &  &  &  & \multicolumn{1}{c}{$\textit{X} = 8$} & \multicolumn{1}{c}{$\textit{X} = 24$} \\
        \cmidrule[0.5pt]{1-7}
        \scriptsize{\textit{V}$256$\textit{-}$6$/\textit{X}} & $256$ & $6$ & $8$ & $5.1$ & $1.22$ & $0.120$ \\
        \scriptsize{\textit{V}$192$\textit{-}$12$/\textit{X}} & $192$ & $12$ & $3$ & $5.6$ & $1.43$ & $0.136$ \\
        \scriptsize{\textit{V}$256$\textit{-}$12$/\textit{X}} & $256$ & $12$ & $4$ & $9.9$ & $2.44$ & $0.240$ \\
        \scriptsize{\textit{V}$384$\textit{-}$12$/\textit{X}} & $384$ & $12$ & $6$ & $21.8$ & $5.25$ & $0.538$ \\
        \scriptsize{\textit{V}$512$\textit{-}$12$/\textit{X}} & $512$ & $12$ & $8$ & $38.6$ & $9.13$ & $0.953$ \\
        \scriptsize{\textit{V}$640$\textit{-}$12$/\textit{X}} & $640$ & $12$ & $10$ & $60.0$ & $14.1$ & $1.49$ \\
        \scriptsize{\textit{V}$768$\textit{-}$12$/\textit{X}} & $768$ & $12$ & $12$ & $86.2$ & $20.1$ & $2.14$ \\
        \bottomrule
    \end{tabulary}
    }
    \vspace{3mm}
    \caption{Details on the ViT models we are training. We use the standard \textit{s}, \textit{S}, \textit{Ti}, \textit{B} model sizes, as well as other intermediate model sizes. To simplify and unify notation, we adopt the naming convention \textit{V}$d$\textit{-}$L$/\textit{X} for a Vision Transformer of depth $L$ and embedding dimension $d$. Here \textit{X} refers to the patch size.}
  \end{subfigure}
  \caption{(Left) Hyperparameters are optimized across model classes. (Right) The ViT models used for this study.}
  \label{fig:models}
  \vspace{-2mm}
\end{figure*}

\paragraph{Fixed patch size training.}
Different patch sizes come at different computational costs; the number of tokens $n$ scales with $\mathcal{O}(1 / p^2)$ and thus processing inputs scales with $\mathcal{O}(1 / p^4)$ due to quadratic dependence on the input sequence length of the self-attention operation\footnote{The exact complexity is $\mathcal{O}(1 / p^4 \times d + 1 / p^2 \times d^2)$ for patch sizes $d > n$ where, unless packing is performed, $\mathcal{O}(1 / p^2 \times d^2)$ is the dominant term.}. Consequently, a reduction in the patch size results in a substantial increase in the computational requirements for a forward pass. We present our empirical analysis in Fig.~\ref{fig:flops_ps}. Using smaller patch sizes is often desirable as it yields superior model performance when paired with enough compute. To explore this trade-off, we pre-train Vision Transformers of different sizes (see Fig.~\ref{fig:models}(b) for a summary) on the public \Imagenet{21} dataset~\citep{ridnik2021imagenet21k}, employing various patch sizes that remain \emph{fixed} throughout training. To enhance computational efficiency and prevent bottlenecks caused by data transfer, we resize images to dimensions where $h = w = 120$, utilizing the FFCV data loader~\citep{leclerc2023ffcv}
\footnote{We expect a small decrease in performance due to this decreased resolution. Our most compute-intensive models (ViT Base variant) achieve top-$1$ accuracy on \Imagenet{1} $79.2$ \% when fine-tuned and $77.2$ \% when linear probed on top of the extracted embeddings. ~\citet{steiner2021train} report $80.42 \%$ fine-tuning performance for a \textit{ViT-B}/$16$ model on $224 \times 224$ images trained for 30 epochs on \Imagenet{21}, which already surpasses our maximum compute budget.}.
This approach enables the application of a variety of patch sizes $p \in \{120, 60, 30, 24, 20, 15, 12, 10, 8, 6, 4, 3, 2, 1\}$, each of which perfectly divides the input resolution. During the training phase, we perform data augmentation, specifically random cropping, and horizontal flipping, and measure the 10-shot error (denoted as $E$) on the \Imagenet{1} dataset~\citep{deng2009imagenet}. This is because upstream performance metrics may not reliably indicate model effectiveness~\citep{tay2022scaling, zhai2022scaling}.
While multiple epochs over the same dataset have been identified as less effective in language modeling tasks~\citep{xue2023repeat, muennighoff2023scaling}, the use of augmentations in this work supports the feasibility of multi-epoch training without significant degradation in performance. This observation holds for the data and computational scales  (up to $10$ EFLOPs) that we consider~\citep{zhai2022scaling}.

When calculating compute $C$, we exclude the computations associated with the `head' of the network that maps the embedding dimension to the number of classes~\citep{kaplan2020scaling}. Additionally, we adopt the approximation of previous work that the FLOPs required for the backward pass are approximately equivalent to twice the FLOPs incurred during the forward pass. Here, we are optimizing for FLOPs, and do not account for different types of hardware accelerators. 
For highly parallel neural architectures, FLOPs in general exhibit a strong correlation with accelerator time (see e.g. Fig. 4 (right) by \citet{alabdulmohsin2023getting} for ViTs specifically and our results in App.~\ref{app:time_measurement}). 
In our study, we focus exclusively on Transformer models which are very hardware-efficient~\citep{dosovitskiy2020image}. More details regarding the experimental setup are provided in App.~\ref{app:experimental_setup}.

For a fixed model shape, we fit power laws for every patch size in terms of compute (which is proportional to the number of examples seen in this case). The power law takes the form\footnote{As aforementioned, our models are bound by data rather than the number of parameters.}
\begin{equation}
\label{eq:single_scaling_law}
E_P = f_P(C) = a_P (C + d_P)^{-b_P} + c_P.
\end{equation}
where the exponent $b_P$ dictates the decay speed and $c_P$ corresponds to the maximal reachable performance given infinite compute.
After fitting the parameters $a_P, d_P, b_P, c_P \in \mathbb{R}^+$, we can predict downstream performance $E_P$ (\Imagenet{1} 10-shot top-$1$ unless otherwise stated) as a function of compute $C$ measured in FLOPs.
We display the results for the \textit{V}$640$\textit{-}$12$ model in Fig.~\ref{fig:patch_size_640_12} and Fig.~\ref{fig:patch_size_640_12_scheduler}. 
We provide analogous plots for all model sizes in App.~\ref{app:additional_results}. 
From those scaling trends, it is evident that different patch sizes are optimal for different amounts of compute. 

In other words, \emph{given the same compute, different patch sizes yield different improvements at specific levels of performance.}
Given that insight, a very natural question emerges:
\begin{center}
\textit{Can we traverse between the scaling laws more efficiently by allowing for adaptive patch sizes?}
\end{center}

\section{Adaptive Patch Sizes and Traversing Scaling Laws}
\label{sec:traversing}

\paragraph{Adaptive patch size.} To allow for a smooth traversal of different laws, we first need a mechanism that enables mapping a ViT $f_P$ with patch size $P$ to a ViT $f_Q$ with patch size $Q$, while ideally not degrading performance, i.e. $f_P \approx f_Q$. \flexivit~introduced by \citet{beyer2023flexivit} achieves this. It redefines both the patch embedding $\bm{W}_{emb}$ and the positional encodings $\bm{W}_{pos}$ for a fixed base patch size. In every forward pass, depending on the patch size, the base embedding parameters $\bm{W}_{emb}$ are resized based on the pseudo inverse of the resizing matrix. Similarly, the base positional encodings $\bm{W}_{pos}$ are bi-linearly interpolated, enabling the model to change the patch size without a strong performance degradation. For further information, we direct readers to the work by \citet{beyer2023flexivit}.
\begin{figure*}[!t]
    \centering
    \includegraphics[width=1\linewidth]{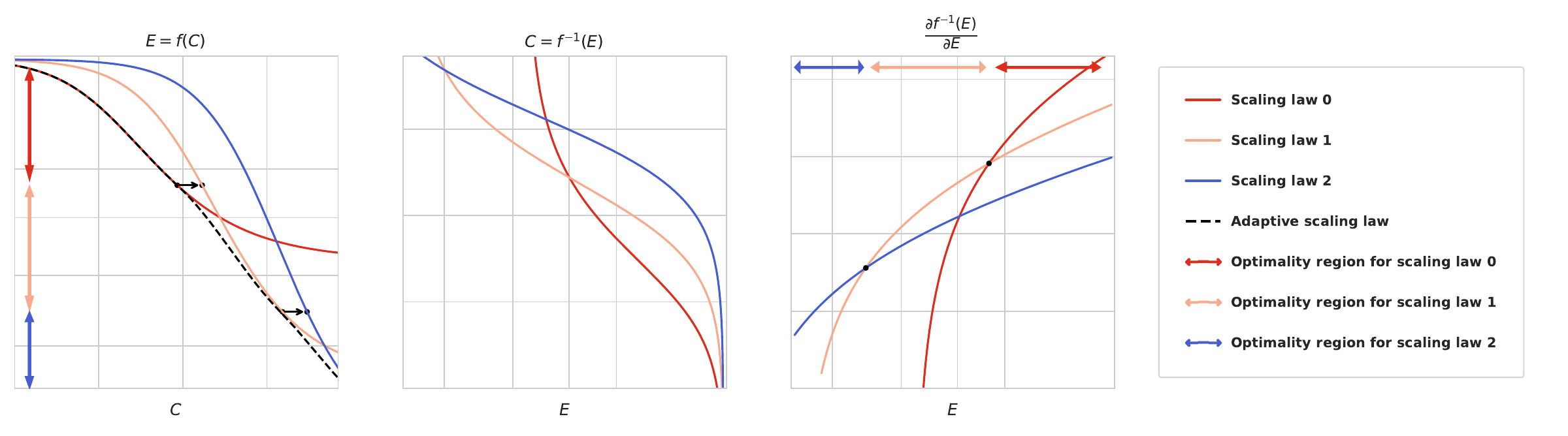}
    \caption{(Left) Different scaling law curves (function $f$ in Eq.~\ref{eq:single_scaling_law}) corresponding to different training configurations. Black arrows indicate points of transition between scaling laws. (Middle) We illustrate the inverse of the above function $f^{-1}$ for the same scaling law curves. (Right) We visualize the gradient of the inverse $\partial f^{-1}(E) / \partial E$ for the same scaling laws. Taking the curve that maximizes the aforementioned gradient, leads to a partition of the space. From this partition, we can deduce a strategy determining which scaling law to `follow' for each performance level.}
    \label{fig:scaling_laws_viz}
\end{figure*}
\vspace{-2mm}
\paragraph{Traversing scaling laws.} Let the set of scaling laws be $\{f_P\}$ with $P$ denoting the patch size. Each law maps a given level of compute $C$ to the predicted downstream performance $E_P=f_P(C)$. Consider the inverted laws $f_P^{-1}(E)$ which predict for a given level of desired performance $E$, how much compute $C$ needs to be invested. 
For a given error level $E^{*}$, we aim to reach a lower error $E^{*}-\eps$ for $\eps > 0$, while spending the minimal amount of compute $C$ to achieve this, i.e. we want the least change in $f_P^{-1}$.
To solve this problem, we simply compute the partial derivatives
\begin{equation}
\label{eq:grad_equation}
    q_P(E^{*}):=\frac{\partial f_P^{-1}(E)}{\partial E}\Big{|}_{E=E^{*}} \hspace{2mm} \forall P.
\end{equation}
Maximising $q_P$ over the patch size $P$, partitions the error space disjointly (e.g. if we assume $E$ is the classification error taking values in $[0,1]$), 
\begin{equation*}
    [0, 1] := \bigcup_{P}E_P,
\end{equation*}
where $E_P \subset [0,1]$ denotes the set where the patch size $P$ achieves the most efficient improvement. This partition
naturally gives rise to a scheduler for the patch size, which empirically turns out to be monotonic (i.e. starting from the largest patch size for large classification error values and ending with the smallest for small classification errors), which is expected based on the observations in Fig.~\ref{fig:patch_size_640_12}. We visualize the strategy in Fig.~\ref{fig:scaling_laws_viz}.

\begin{minipage}{.49\textwidth}
  \centering
  \includegraphics[width=1\linewidth]{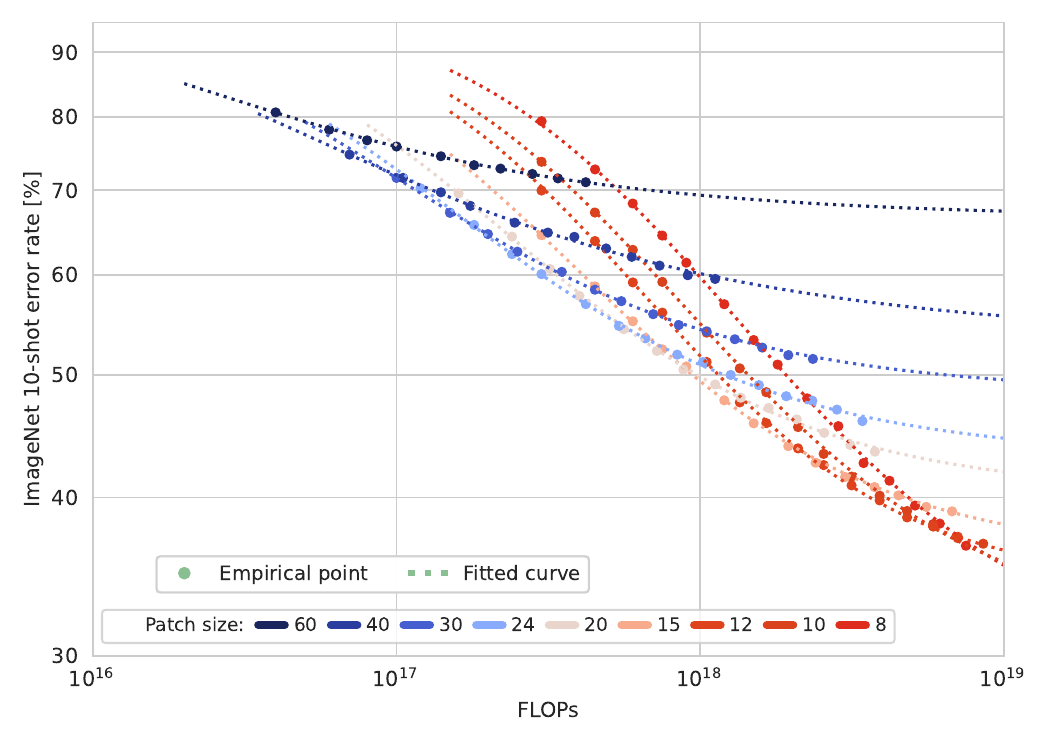}
  \captionof{figure}{Downstream performance as a function of compute for the \textit{V}$640$\textit{-}$12$ model and different patch sizes. We use a log-log scale.}
  \label{fig:patch_size_640_12}
\end{minipage}
\hfill
\begin{minipage}{.49\textwidth}
  \centering
  \includegraphics[width=1\linewidth]{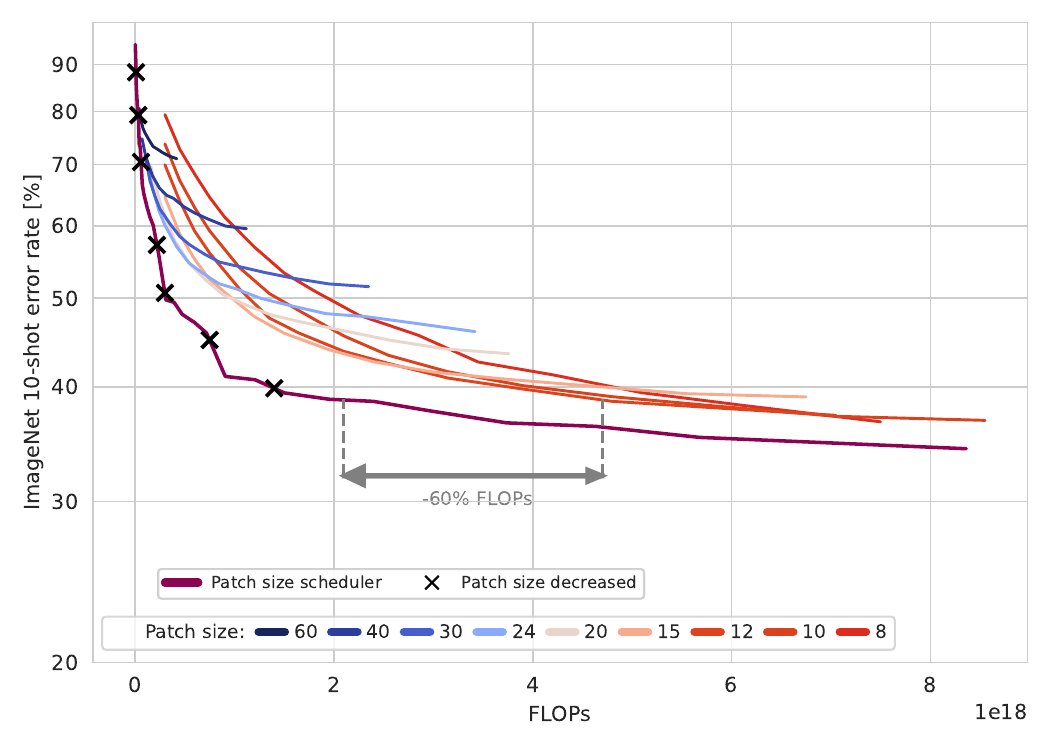}
  \captionof{figure}{Downstream performance of the \textit{V}$640$\textit{-}$12$ trained with our patch size scheduler, and its potential benefits.}
  \label{fig:patch_size_640_12_scheduler}
\end{minipage}

\paragraph{Scheduled training.} We now test the devised strategy in a practical setting by pre-training variously-sized ViTs on \Imagenet{21} using our patch size scheduler. We use the same training setup as for the fixed patch size experiments and let the scheduler consider patch sizes $P \in \{60, 40, 30, 24, 20, 15, 12, 10, 8\}$. We display \Imagenet{1} 10-shot error rate as a function of compute $C$ for the model \textit{V}$640$\textit{-}$12$ in Fig.~\ref{fig:patch_size_640_12_scheduler} and provide plots for all other models in the App.~\ref{app:additional_results}. The crosses denote the points where the scheduler switches patch size. We observe a significant improvement in terms of compute efficiency, allowing for up to $-60\%$ FLOPs to achieve the same performance. 
While switching patch sizes may initially result in a slight decrease, attributed to changes in the entropy within the self-attention layers, this minor setback is rapidly offset as the image is parsed in a more fine-grained manner.
Such degradation is thus not even visible in Fig.~\ref{fig:patch_size_640_12_scheduler}\footnote{Differences in the effective receptive field of each patch are typically mitigated by cropping as a component of the training procedure.}. To facilitate comparison across all model sizes at once, we further visualize the compute-optimal barrier for both fixed and scheduled training in Fig.~\ref{fig:optimality}. By compute-optimal, we refer to a model that optimally trades off model size, patch size, and number of samples seen for a given level of compute $C$, i.e. achieving the lowest error $E$. We observe that the optimally scheduled models significantly outperform the optimal static models, halving the required compute to optimally train a ViT-Base model (for our compute budget).

\begin{minipage}{.49\textwidth}
    \centering
    \includegraphics[width=1\linewidth]{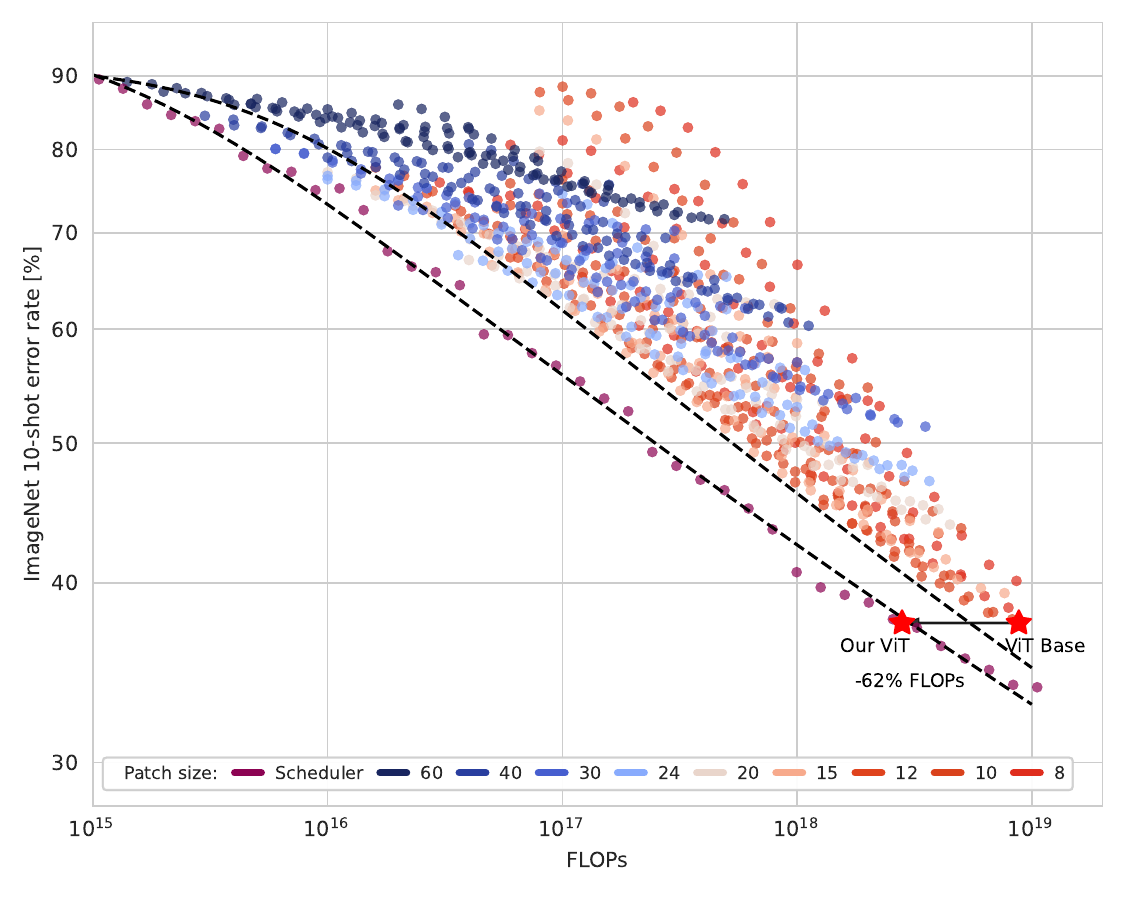}
    \captionof{figure}{Compute-optimal static and scheduled models for various patch and model sizes. We plot using a log-log scale.}
    \label{fig:optimality}
\end{minipage}
\hfill
\begin{minipage}{.49\textwidth}
    \centering
    \includegraphics[width=1\linewidth]{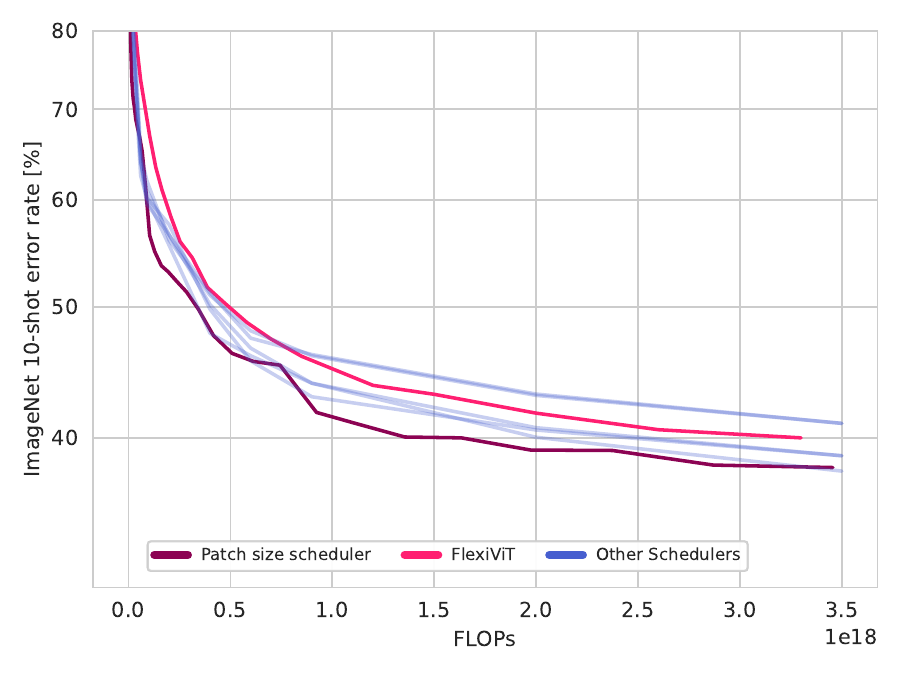}
    \captionof{figure}{We compare performance as a function of training compute against the scheduler of \flexivit~and other schedulers. Irrespective of the current patch size in the scheduler, we use the smallest patch size (i.e. $8$), when evaluating \flexivit.}
    \label{fig:baselines}
\end{minipage}

\paragraph{Is the schedule optimal?} While our scheduler improves over the individual models, it is not clear yet that it does so in an optimal sense, i.e. can other schedules achieve similar benefits? \citet{beyer2023flexivit} also employ a patch size scheduler but use a uniformly random sampling of the patch size at every step. We compare against their model \flexivit~in Fig.~\ref{fig:baselines} and observe that our scheduler indeed outperforms~\flexivit as expected; \flexivit~targets a lower inference cost by making the model robust to many patch sizes (hence the random scheduler). Compute-optimality is not their objective. 
Additionally, we conduct comparisons with straightforward patch size scheduling strategies—both linear and logarithmic. Specifically, for a predetermined total computational budget, we distribute the transition points evenly or according to a logarithmic scale across the training process.
This way, we assess whether simply any monotonic scheduler leads to the same improvements, or whether the position of the transition points matter. We display the results in Fig.~\ref{fig:baselines}. We again observe that our scheduler remains more efficient, carefully determining the transition points based on the scaling laws, thus indeed leading to a significant improvement.

\begin{wrapfigure}{r}{0.5\textwidth}
    \centering
    \includegraphics[width=1\linewidth]{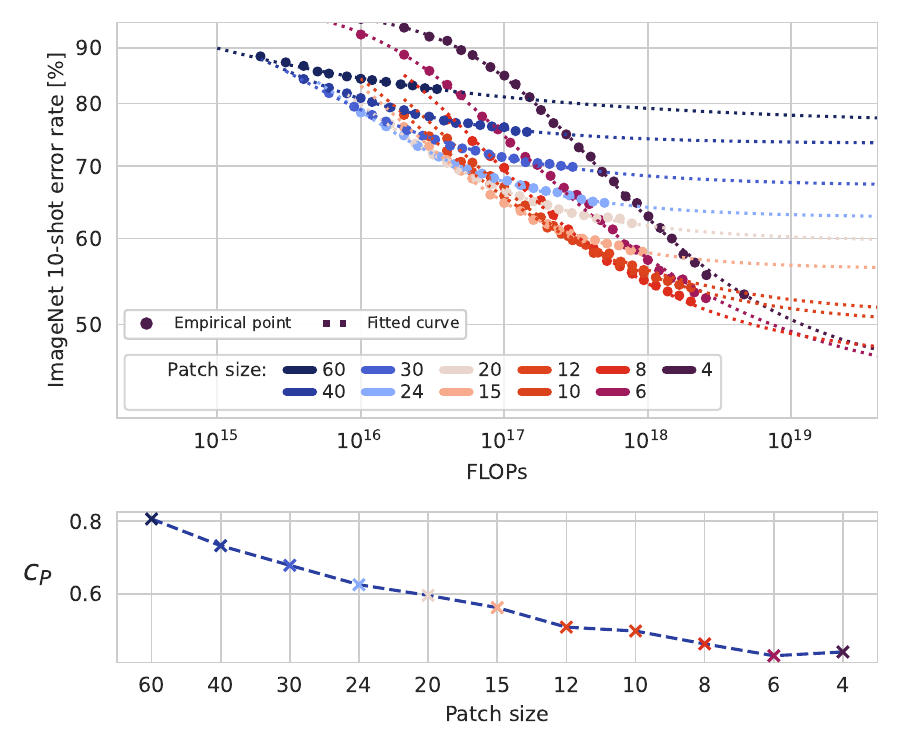}
    \caption{We train the \textit{V}$256$\textit{-}$6$ with smaller patch sizes. This does not lead to a monotonically better performance.}
    \label{fig:smaller-patch}
    \vspace{-8mm}
\end{wrapfigure}

\paragraph{Smaller patch sizes.} Undeniably, the choice of patch size affects the inductive bias of ViTs --- in general, the mechanism of `tokenization' in the input affects the inductive bias of any Transformer model --- by controlling the amount of compute and the level of details we are interested in extracting from an image. The patch size also controls the overall sequence length $n$ processed by the Transformer model, and therefore the degree of weight sharing between the parameters. Our previous laws clearly show that smaller patch sizes lead to better performance in high-compute areas. But does this trend also extend to even smaller patch sizes? We explore this question empirically by using the same experimental setup and pre-training on even smaller patch sizes $P \in \{6, 4\}$ in addition to the previous results. We display the results in Fig.~\ref{fig:smaller-patch}. 

We observe that while some absolute gains in performance can still be achieved with patch size $6$, the additional required amount of compute is extremely high. For the even smaller patch size $4$, one actually starts to lose in performance, as can be seen from plotting the intercepts $c_P$ of the corresponding scaling laws. 
The behavior of performance with respect to patch size is thus only monotonic up to a certain point, and performance may actually worsen beyond that. This is in contrast to other scaling parameters such as the number of samples or the model size that usually offer a monotonic behavior in performance when scaled appropriately.

\section{Adaptive Context Size of an LLM}
\label{sec:other_domains}

While the previous chapter focused on Vision Transformers, the aforementioned intuitions and results generalize also to other domains. In this section we present a compelling case for the training of a Transformer-based language model.
A critical consideration in training such a model is determining the optimal context size, as most Transformer models do not inherently support context extrapolation out of the box~\citep{kazemnejad2023impact}.
The context size determines the amount of information the model can use when making predictions, with an expanded context being essential for enhanced performance in certain downstream tasks~\citep{dao2022flashattention, tay2020long}. A larger context size, however, comes with increased computational requirements, similar to the patch size in ViTs. 

To mitigate the significant training overhead associated with long contexts --- where the quadratic complexity of self-attention becomes a bottleneck --- a common strategy involves initially training with shorter contexts. Subsequently, the model is fine-tuned on longer contexts, employing a modest computational budget to balance performance and efficiency~\citep{chen2023longlora}.
Here, we demonstrate how different context sizes lead to different learning improvements at different scales of compute, and navigating across the optimal scaling law can lead to \emph{substantial performance gains} given the same amount of compute. 
We train Transformer models~\citep{touvron2023llama, touvron2023llama2} on English Wikpedia and Books (more details in App.~\ref{app:other_domains}).

\begin{figure}[h]
    \centering
    \includegraphics[width=0.7\linewidth]{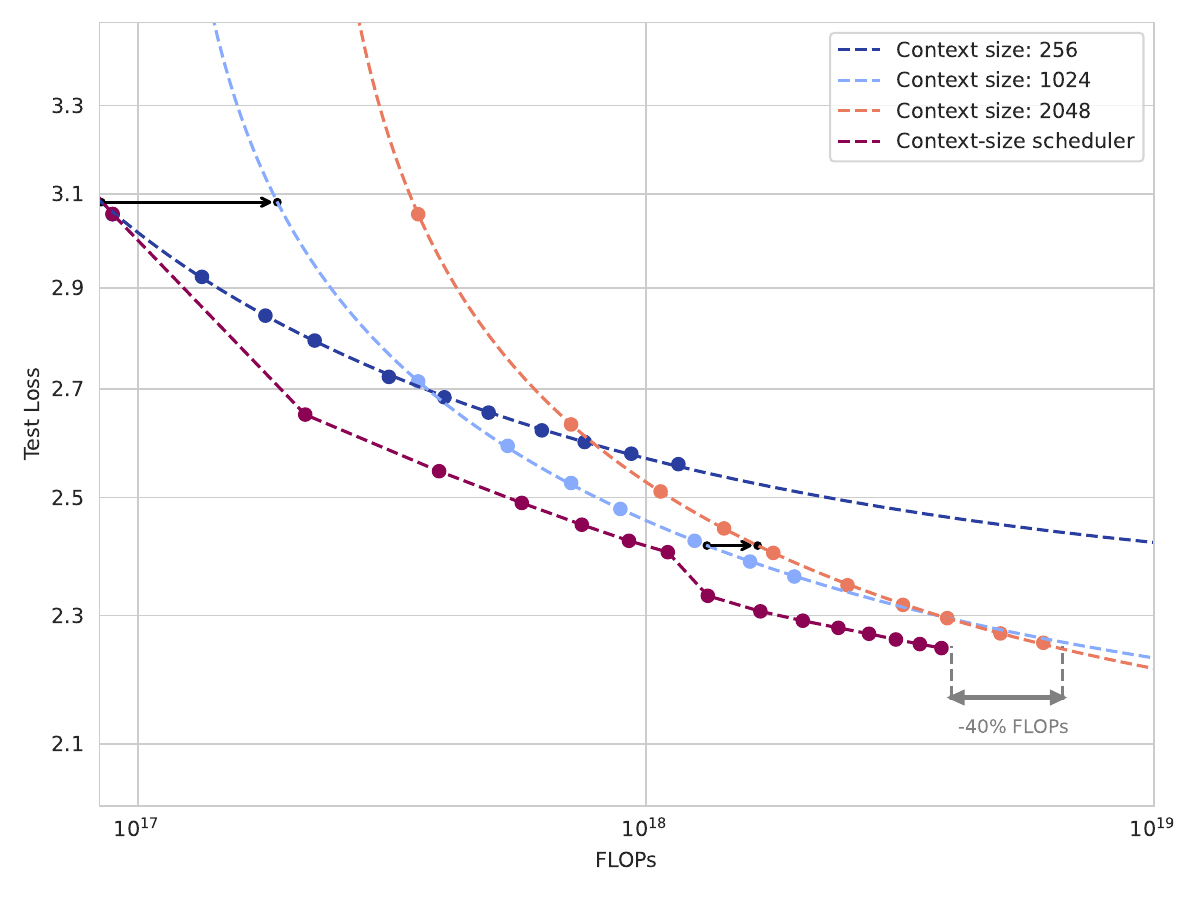}
    \caption{We compare model performance throughout training with dynamic (magenta) and fixed context sizes.}
    \label{fig:llama_context_size}
\end{figure}

The findings illustrated in Fig.~\ref{fig:llama_context_size} align with our hypothesis: smaller context sizes prove to be more optimal at the beginning of the training process, while larger contexts yield greater efficiency as training progresses. Employing a similar approach to that used with patch sizes, we introduce a scheduler that is adjusted in accordance with scaling laws. This strategy, akin to our adjustments in patch size, results in substantial computational savings, enabling a reduction in FLOPs of up to $40\%$.

\section{Other Shape Parameters}

To further verify the efficiency of our approach, we study different `shape' parameters when training a Vision Transformer. 

\subsection{Adapting Model Width}
\label{sec:adapting-width}

\begin{figure}[t]
\centering
\begin{minipage}{.375\textwidth}
  \centering
  \includegraphics[width=1\linewidth]{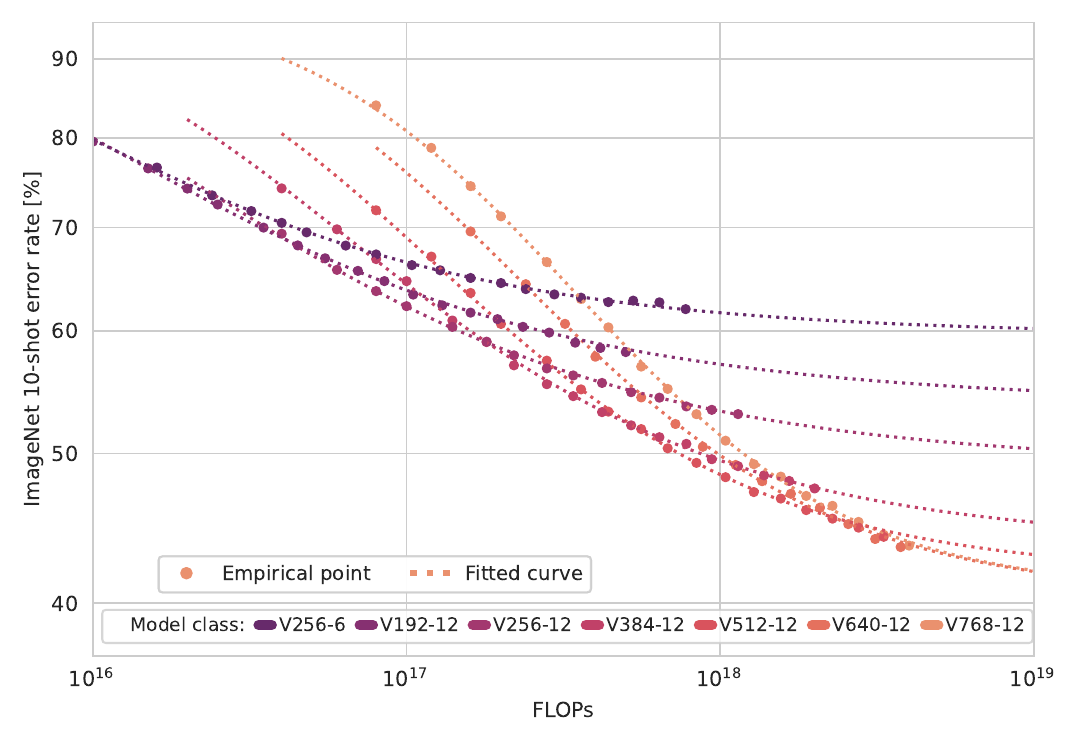}
  \vspace{-7mm}
  \captionof{figure}{Downstream performance as a function of compute for ViT of different size, trained with a patch size of 20. We use a log-log scale.}
  \label{fig:model_class_20}
\vspace{0mm}
\end{minipage}%
\hspace{3mm}
\begin{minipage}{.59\textwidth}
  \centering
  \includegraphics[width=1\linewidth]{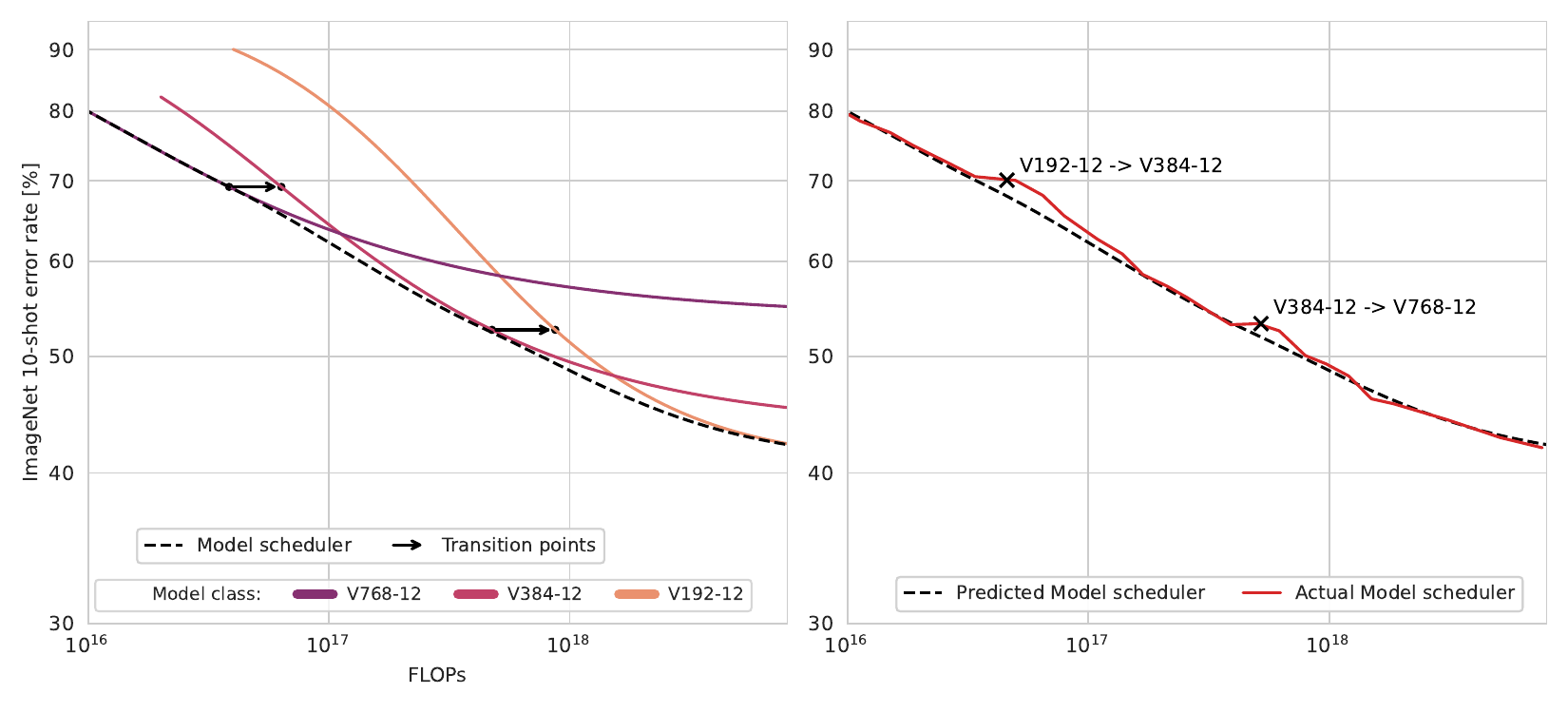}
  \vspace{-8mm}
  \captionof{figure}{The theoretically predicted scheduled performance (left) and the empirical obtained (right) performance. While transitions are less smooth, the model based on the scheduler quickly recovers back to the predicted law.}
  \label{fig:ms_scheduler_20}
\vspace{0mm}
\end{minipage}
\end{figure}

Similarly to the patch size, we need a mechanism that maps a transformer of smaller width $d_1$ to a transformer of larger width $d_2$. This is a very well-studied problem, especially in the case of natural language processing, and many schemes have been proposed~\citep{gesmundo2023composable, chen2022auto, gong2019efficient, yao20232x, wang2023learning, lee2022towards, shen2022staged, li2022automated}. Here, we focus on the simplest approach, where we expand the initial model $d_1$ by adding randomly initialized weights (see App.~\ref{app:additional_results} for details).  Although our expansion is not function preserving --- i.e. we can expect a small drop in performance immediately after adapting the model (see Fig.~\ref{fig:ms_scheduler_20}) --- we found that the model quickly recovers, and hence conclude that while not ideal, this simple expansion mechanism suffices for our setting\footnote{In practice we found that proposed function preserving schemes that depend on zero-initializing weights in the network, e.g. by ~\citet{gesmundo2023composable}, perform suboptimally.}.

\paragraph{Scaling width.} 
The role of the model width and its associated scaling properties have been long analysed in the literature~\citep{zhai2022scaling, alabdulmohsin2023getting}. We repeat the scaling study for our own experimental setup and pre-train Vision Transformers of various widths and training durations on \Imagenet{21}, using the same experimental setup as detailed in Sec.~\ref{sec:patch-size}. In Fig.~\ref{fig:model_class_20} we report 10-shot \Imagenet{1} error as a function of compute for a fixed patch size $P=20$ (more details and results for other patch sizes are provided in the App.~\ref{app:additional_results}). We again observe that different model widths are optimal for different levels of compute, similarly offering the potential for computational speed-ups by adapting the shape throughout training. 

\paragraph{Scheduling width.} It is worth noting that strategies for expanding models during training have been previously explored. However, the critical question of \textit{when} this expansion should occur has largely remained unanswered. Our approach then offers a straightforward and principled solution. We consider three width settings $d \in \{192, 384, 768\}$ and devise our scheduler based on the scaling law as outlined in Sec.~\ref{sec:traversing}. We display the obtained optimal schedule and the actual resulting performance in Fig.~\ref{fig:ms_scheduler_20}. As remarked previously, changing the model width does lead to a momentary deterioration of the performance, but it smoothly recovers back to the predicted performance. We again observe that the scheduled model remains optimal throughout training when compared to the static models. Note that adapting multiple shape parameters --- namely both patch size and model size --- is possible and leads to further improvements, as we showcase in App.~\ref{app:model_patch_together}.

\subsection{Adapting the Training Objective}
\label{sec:other_parameters}

In previous experiments, we have fixed training hyperparameters and focused on adapting the model's shape. Training hyperparameters can nonetheless also be altered optimally. Previous work has already established that bigger \emph{batch sizes} are beneficial during later stages in training~\citep{kaplan2020scaling, hoffmann2022training, zhai2023sigmoid}. Different \emph{training objectives} are also known to contribute differently to downstream performance at distinct stages in training~\citep{zhai2023sigmoid, singh2023effectiveness}. Previous work has relied on heuristics and brute force exploration to determine when these should change. Here, we demonstrate how scaling laws can help decide when to change these parameters and descend based on the optimal one.

\begin{figure}[t]
    \centering
    \includegraphics[width=0.49\linewidth]{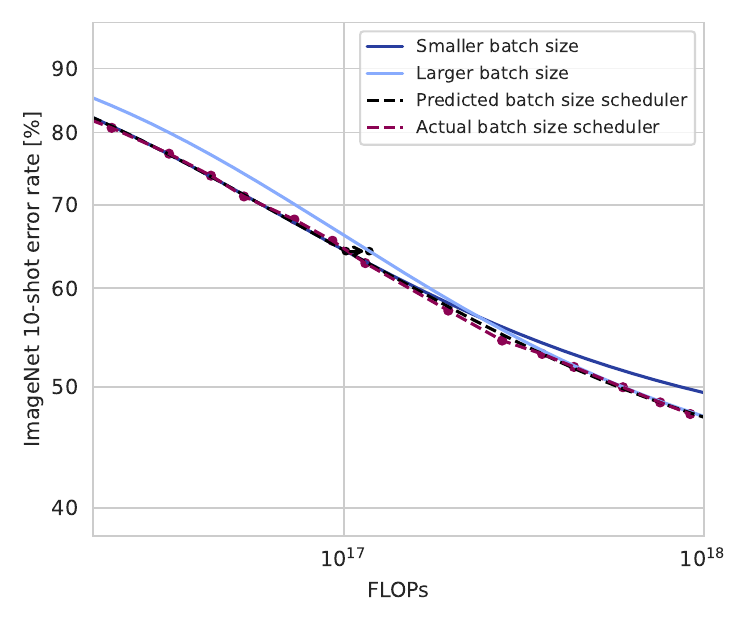}
    \includegraphics[width=0.49\linewidth]{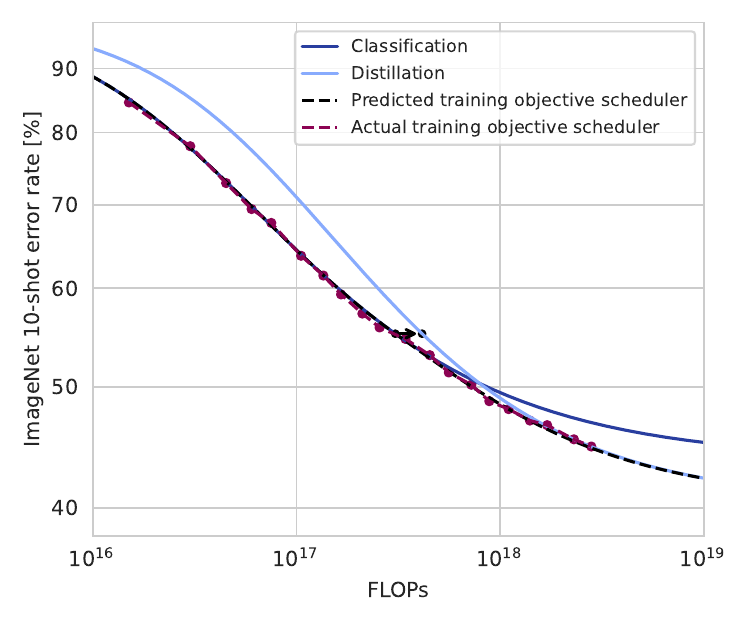}
    \vspace{-5mm}
    \caption{(Left) ViTs with small and big batch sizes and using (right) different upstream objectives. Different optimality regimes are observed for the different settings. Changing between them leads to optimal performance gain for a fixed compute budget.}
    \label{fig:extra_parameters_scheduler}
    \vspace{-4mm}
\end{figure}

\paragraph{Batch size.} We first focus on the batch size used during training. We train two Vision Transformers, one at a larger batch size and the other at a smaller batch size. We find that in terms of number of FLOPs, the smaller batch size initially dominates but again is surpassed at later stages in training by the large batch size run (Fig.~\ref{fig:extra_parameters_scheduler} left). 
Our strategy allows us to maximally take advantage of this difference by optimally transitioning between the two batch sizes, leading to a more optimal model. 

\paragraph{Distillation.} We additionally train ViT models by distilling from a powerful teacher, a bigger pre-trained ViT (Fig.~\ref{fig:extra_parameters_scheduler} right). Distilled labels naturally come at an additional computational cost, due to queries to the teacher, and lead to slower convergence in terms of FLOPs initially in training (FLOPs here include the teacher compute). Distillation objectives, however, were found to lead to increases in performance \citep{hinton2015distilling, pmlr-v80-furlanello18a, pmlr-v139-touvron21a}. Thus, in later stages of training, such an objective will dominate the standard supervised loss. We again leverage this discrepancy and optimally switch from the standard to the distilled loss, allowing us to reach the same level of performance with fewer FLOPs.

\section{Conclusion}
We have explored strategies to train models with variable shape parameters throughout the training process. 
By efficiently traversing scaling laws, we have illustrated the optimal scheduling of shape parameters --- including patch size, context length, model width, and other hyperparameters --- yielding notable computational efficiency for a given performance level.
We further observe that models with dynamically scheduled shape parameters consistently outperform their static counterparts in terms of computational efficiency during training. This underscores the effectiveness of our method.
Our scheduling approach is highly adaptable and applies to any shape parameter that allows for a smooth transition between models of varying configurations.
This opens up many opportunities for future research, applying our scheduling method to other shape parameters such as model depth, sparsity, or a combination of multiple parameters.
Given the increased computational demand for deep learning, we believe our findings make a crucial contribution to the field.

\section*{Impact Statement}

Our method provides insights into the growing challenges associated with the exponential scaling of compute resources for the training of frontier models. By making the training of large models more accessible, our approach opens doors to a broader audience of researchers and practitioners, fostering innovation and breakthroughs in artificial intelligence. Importantly, this alternative strategy contributes to a reduction in environmental impact, showcasing a commitment to sustainable and responsible advancements in the field. In App.~\ref{app:enviromental_impact} we provide concrete insights on the expected reduction in $CO_2$ emissions for our experimental setup.

\bibliography{main}
\bibliographystyle{iclr2024_conference}

\newpage
\appendix

\section{Limitations}
We detail the limitations of our work to the best of our knowledge.
\begin{itemize}
    \item We used greedy search with a small compute budget to get optimal hyper-parameters per model class. In practice, optimal parameters can change if larger levels of compute are available, as also hinted in Sec.~\ref{sec:other_parameters}. 
    \item In order to determine the optimal scheduler for a given shape parameter, knowledge of its scaling behavior is needed, which comes at a high computational cost. On the other hand, the scaling behavior of many shape parameters has already been established (e.g. width, depth, MLP-dimension \citep{alabdulmohsin2023getting}) and can readily be used in our scheduler.
    \item Accurately predicting compute-optimal models, requires one to accurately schedule the learning rate throughout training. As we are interested in what happens during training for low-budgets of computes we do not schedule the learning rate nor embark on a cooldown phase~\citep{zhai2022scaling}, as this would constitute a large fraction of the overall training time. We expect learning rate schedulers may shift our conclusion but not the outcome and takeaway message.
    \item While we observe that the scheduled models are compute-optimal throughout all of training, we observe the largest gains earlier on throughout training. Indeed, we do not expect our scheduled models to reach better performance for an infinite amount of compute.
\end{itemize}

\section{Experimental Setup}
\label{app:experimental_setup}

We provide more details on the basis on which the experiments were conducted. 

\subsection{Training Details}

\begin{table}[!h]
\vskip 0.15in
\begin{center}
\begin{small}
\begin{sc}
\begin{tabular}{lc}
\toprule
Parameter & Value \\
\midrule
Optimizer & AdamW \\
Betas & $(0.9, 0.999)$ \\
Label smoothing & $0.2$ \\
Weight-decay head & $0.01$ \\
Weight-decay body & $0.01$ \\
Warm-up & 1000 steps \\
Clip gradients' norm & $1.0$ \\
Underlying patch-size shape & $12$ \\
Underlying posemb shape & $8$ \\
\bottomrule
\end{tabular}
\end{sc}
\end{small}
\end{center}
\caption{Hyper-parameters during training. `Underlying patch-size' and `Underlying posemb shape' refer to the flexible modules when training under a flexible patch size scheduler.}
\label{tab:training-details}
\end{table}

\begin{figure}[!h]
    \centering
    \includegraphics[width=0.6\linewidth]{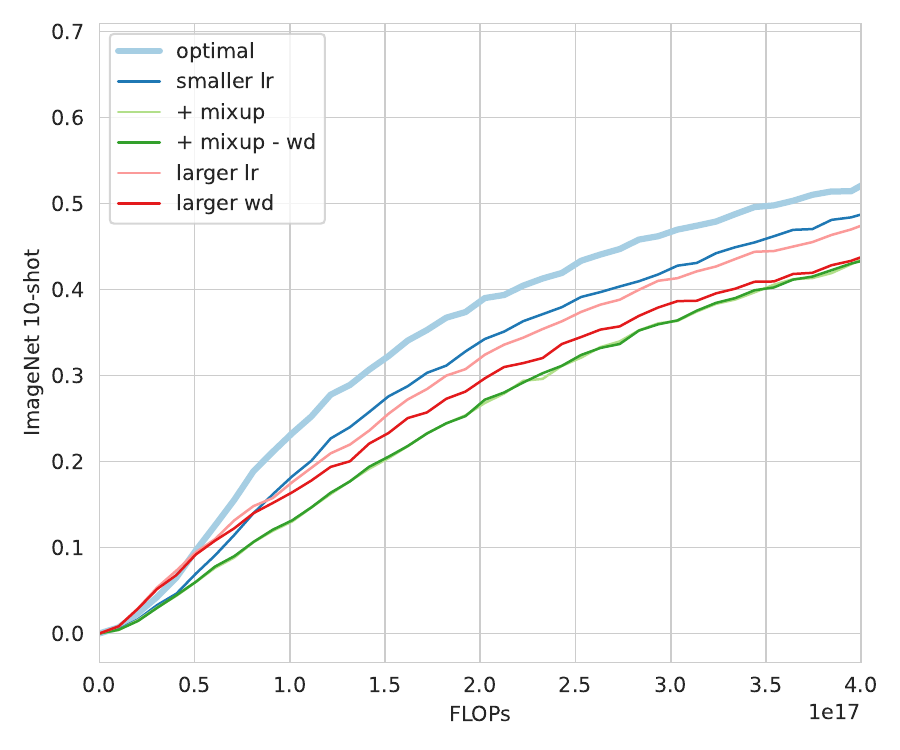}
    \caption{Hyper-parameter search for a fixed (and small) budget of compute.}
    \label{fig:more-ablations}
\end{figure}

In Tab.~\ref{tab:training-details} we showcase hyper-parameters used when training on \Imagenet{21}. We optimized each of the parameters for the different model classes by training for different configurations for a fixed, small amount of compute, namely $4 \times 10^{17}$ FLOPs. Some examples of such hyper-parameter search are illustrated in Fig~\ref{fig:more-ablations}. All experiments were conducted using \textit{bfloat16}.

\subsection{Finetuning Details}

In Tab.~\ref{tab:finetuning-details} we showcase hyper-parameters used when finetuning on \Imagenet{1}. For the few-shot results, we use the \textit{linear\_model.Ridge} function from \textit{scikit-learn} with a regularization parameter of $1e^{-2}$.

\begin{table}[!h]
\vskip 0.15in
\begin{center}
\begin{small}
\begin{sc}
\begin{tabular}{lc}
\toprule
Parameter & Value \\
\midrule
Optimizer & SGD \\
Learning rate & 0.03 \\
Momentum & 0.9 \\
Weight decay & 0.0 \\
Number of steps & 20000 \\
Clip gradients' norm & 1.0 \\
Scheduler & Cosine \\
\bottomrule
\end{tabular}
\end{sc}
\end{small}
\end{center}
\caption{Hyper-parameters during fine-tuning on \Imagenet{1}.}
\label{tab:finetuning-details}
\end{table}

\subsection{Dataset Description}

For the ViT experiments, we follow the protocol of~\citet{ridnik2021imagenet21k} to preprocess \Imagenet{21}. It consists of roughly 12 million images and 11 thousand different classes. This is still considerably lower than the $\geq 29,593$ classes in the \textit{JFT-3B} dataset. We experimented using different weight decay values for the body and the head, as proposed in~\citet{zhai2022scaling} but found no significant difference. We attribute this to the lower number of classes in the dataset we are training in and our value of label smoothing.

\subsection{Scaling Laws}

We fit functions of the form 
\begin{equation}
E =  a (C + d)^{-b} + c.
\end{equation}
Similar to previous work, we resample points to be almost equidistant in the log-domain, in terms of FLOPs. We minimize different initializations using the \textit{minimize} function in \textit{scipy}~\citep{virtanen2020scipy}, and choose the one that leads to the smallest error. The function to minimize is based on the \textit{Huber loss} with $\delta=1e^{-3}$. 

\subsection{Sec.~\ref{sec:other_domains} Details}
\label{app:other_domains}

In Sec.~\ref{sec:other_domains}, we train Llama~\citep{touvron2023llama, touvron2023llama2} models, following the Transformer++ training recipe, as displayed in Tab.~\ref{tab:llama-training}. Due to compute constraints, we train models with embedding dimension $768$ and depth $12$. We train on a subset of the English Wikipedia \textit{20220301.en} and English \textit{bookcorpus} datasets. As done for the ViT models, we select for each configuration the batch size that leads to the most efficient -- in terms of FLOPs -- convergence.

We evaluate on a held-out validation set. Although the validation samples are the same across different runs, we truncate them to match the context size of the respectively trained model. Models with longer contexts are thus expected to achieve lower test loss due to the enhanced context size (conditioned on longer contexts, subsequent tokens are more predictable). As during inference we are interested in the validation cross-entropy loss over the longer contexts, we do not adjust for this discrepancy.

\begin{table}[!h]
\vskip 0.15in
\begin{center}
\begin{small}
\begin{sc}
\begin{tabular}{lc}
\toprule
Parameter & Value \\
\midrule
Optimizer & AdamW \\
Betas & $(0.9, 0.95)$ \\
Clip gradient's norm & $1.0$ \\
Weight decay & $0.1$ \\
Dropout & 0.0 \\
Warmup & $1000$ \\
Norm & \textit{RMSNorm} \\
Bias & No \\
Peak Learning Rate & GPT-3 values \\
\bottomrule
\end{tabular}
\end{sc}
\end{small}
\end{center}
\caption{Hyper-parameters during training of the language models.}
\label{tab:llama-training}
\end{table}

Extrapolating to longer contexts is a very active area of research with exciting work published recently~\citep{qin2023exploring, ruoss2023randomized, press2021train, peng2023yarn}. In our case, we are training using RoPE positional encodings~\citep{su2024roformer}, which are known to extrapolate easier compared to other ones, such as absolute positional encodings.

The exact number for the most compute-intensive point in Fig.~\ref{fig:llama_context_size} is $17825792000$ tokens. The exact model size including the embedding parameters is $137841408$ parameters. That leads to an approximate token per parameter value of $129.3$, past the optimal "Chinchilla" point, leading thus to model that are more inference efficient.

\subsection{Sec.~\ref{sec:other_parameters} Details}

In Sec.~\ref{sec:other_parameters}, we train ViT models using the same hyperparameters found as described above. For the batch-size experiments, we train \textit{V}$384$\textit{-}$20$/\textit{12} models with batch size in $\{256, 2048\}$ and navigate across the scaling laws corresponding to the same values. For the different training objective experiments, we train a \textit{V}$384$\textit{-}$20$/\textit{12} model using either supervised training or distillation from a powerful teacher. We use as a teacher a pre-trained \textit{V}$640$\textit{-}$10$/\textit{12} model and train using only distillation loss as in~\citet{hinton2015distilling}, with a temperature of $T=2$. When calculating the FLOPs of the single step, we include the FLOPs of the teacher only for the forward pass.

\section{Additional Experiments}
\label{app:additional_results}

\begin{figure*}
\centering
\begin{tabular}{cc}
  \includegraphics[width=0.49\linewidth]{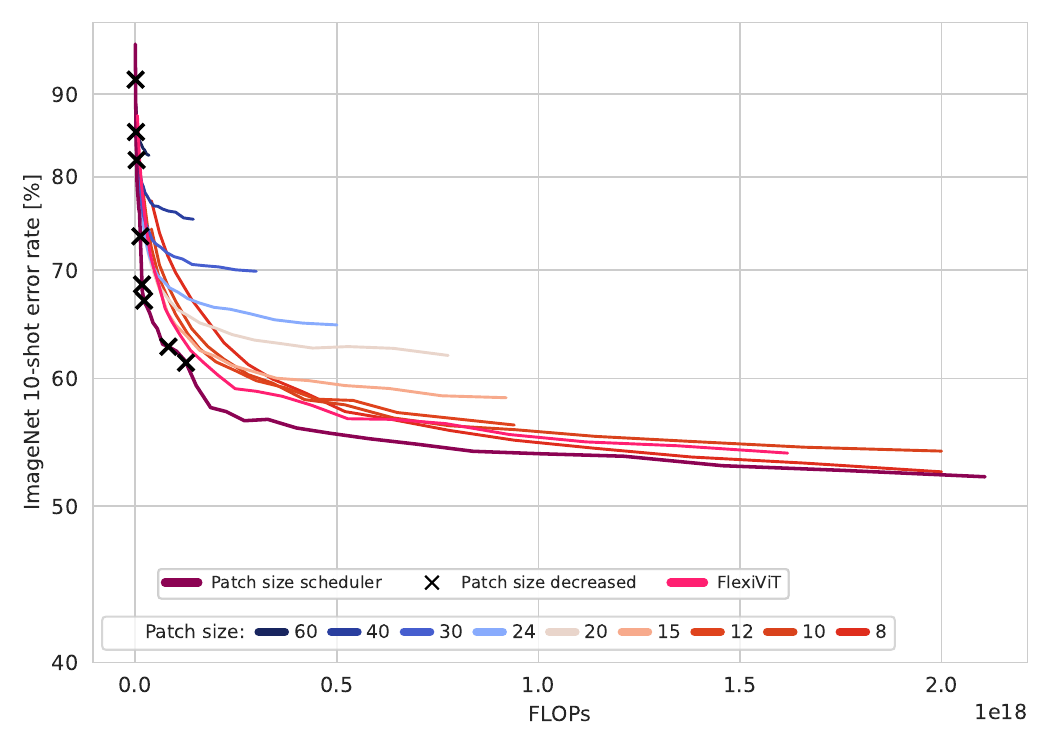} &   \includegraphics[width=0.49\linewidth]{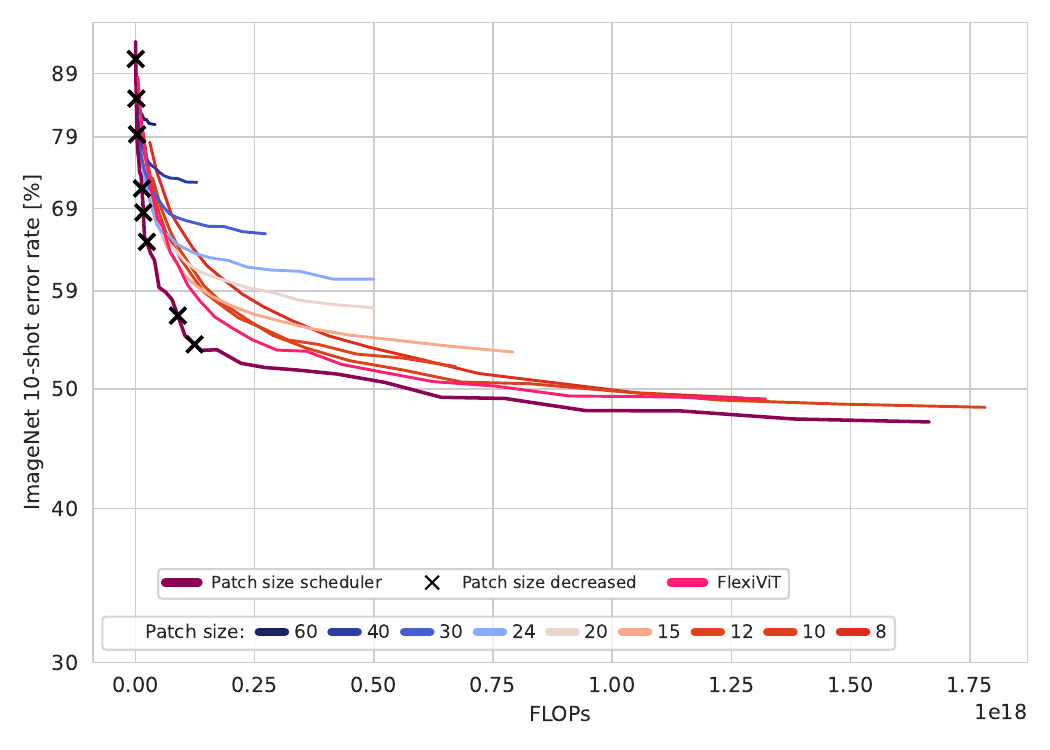} \\
(a) Model \textit{V}$256$\textit{-}$6$. & (b) Model \textit{V}$192$\textit{-}$12$. \\[6pt]
  \includegraphics[width=0.49\linewidth]{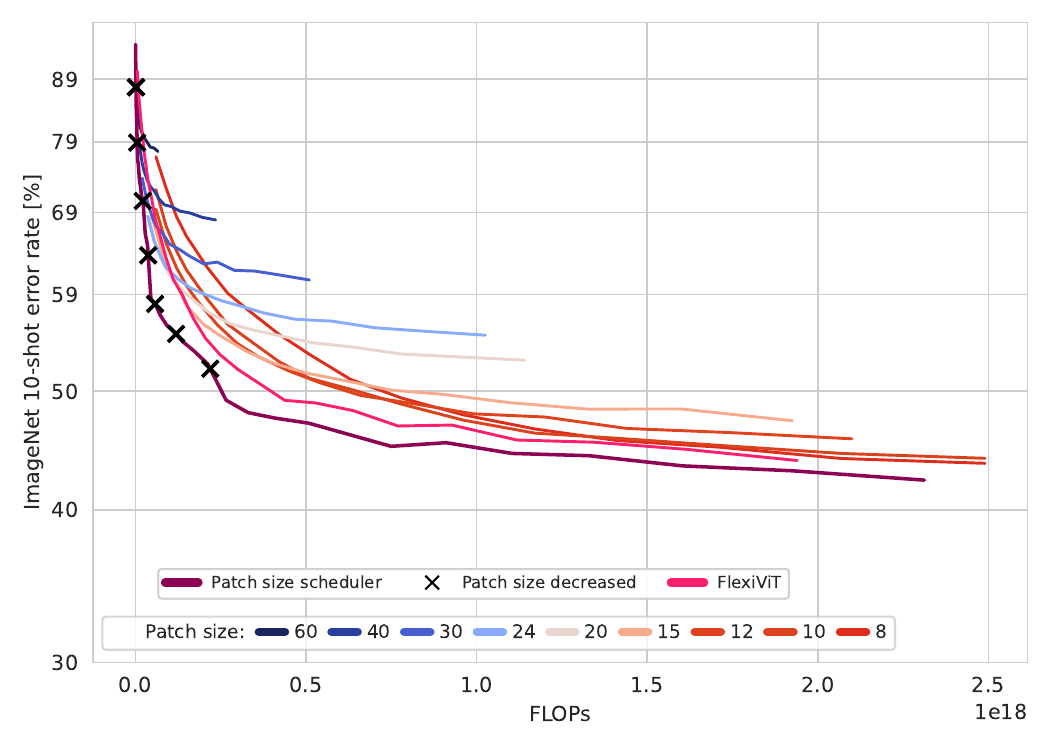} &   \includegraphics[width=0.49\linewidth]{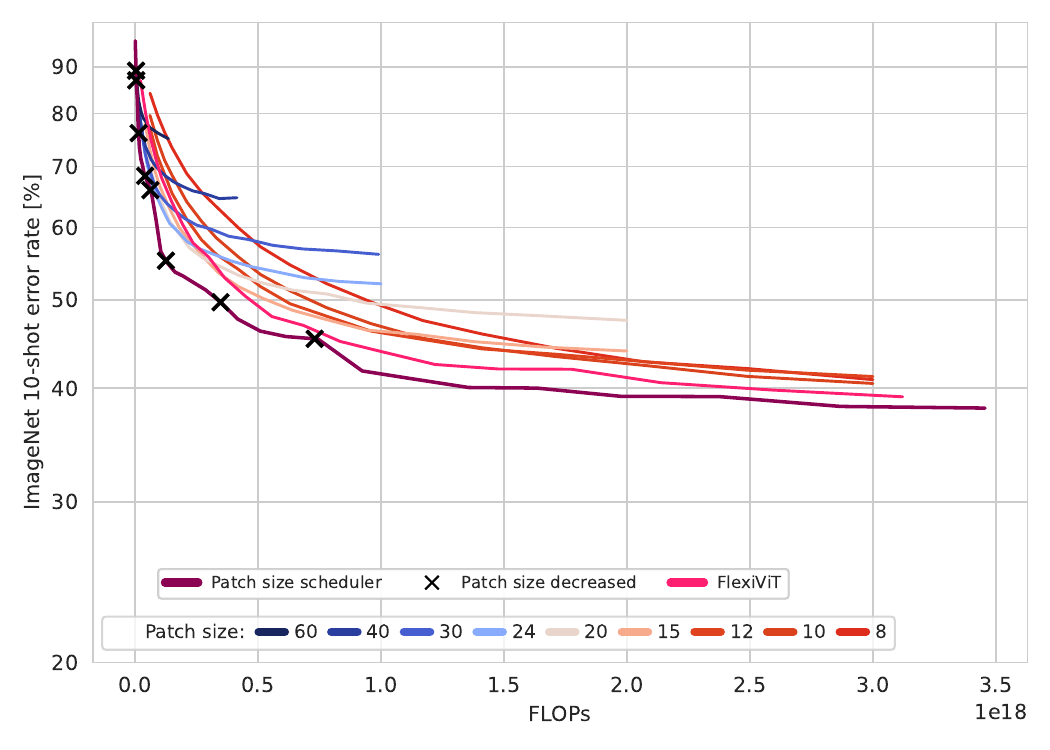} \\
(c) Model \textit{V}$256$\textit{-}$12$. & (d) Model \textit{V}$384$\textit{-}$12$. \\[6pt]
  \includegraphics[width=0.49\linewidth]{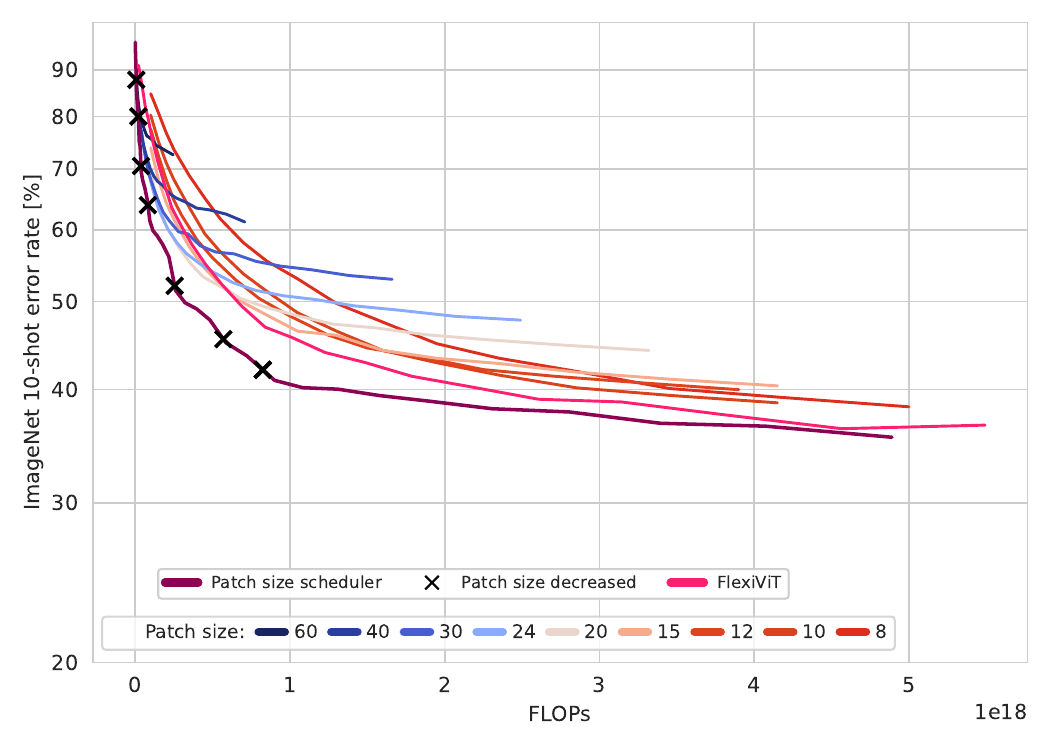} &   \includegraphics[width=0.49\linewidth]{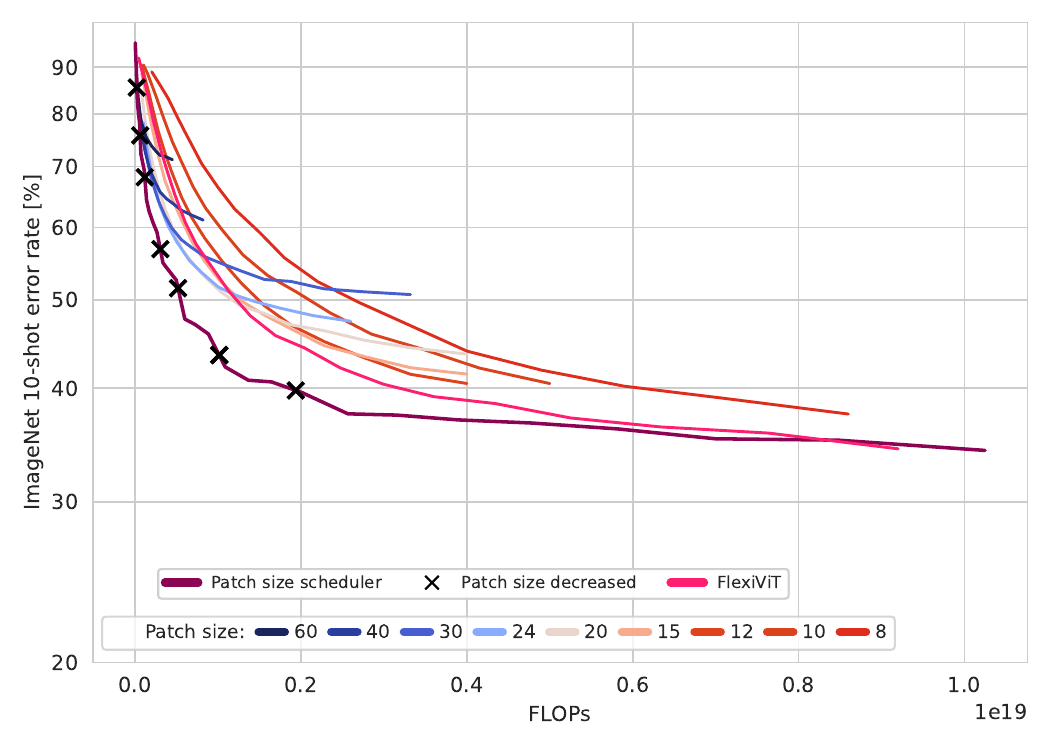} \\
(e) Model \textit{V}$512$\textit{-}$12$. & (f) Model \textit{V}$768$\textit{-}$12$. \\[6pt]
\end{tabular}
\caption{Patch size schedulers for all the remaining model classes analysed.}
\label{ref:patch_size_scheduler_additional}
\end{figure*}

\begin{figure*}
\centering
\begin{tabular}{cc}
  \includegraphics[width=0.48\linewidth]{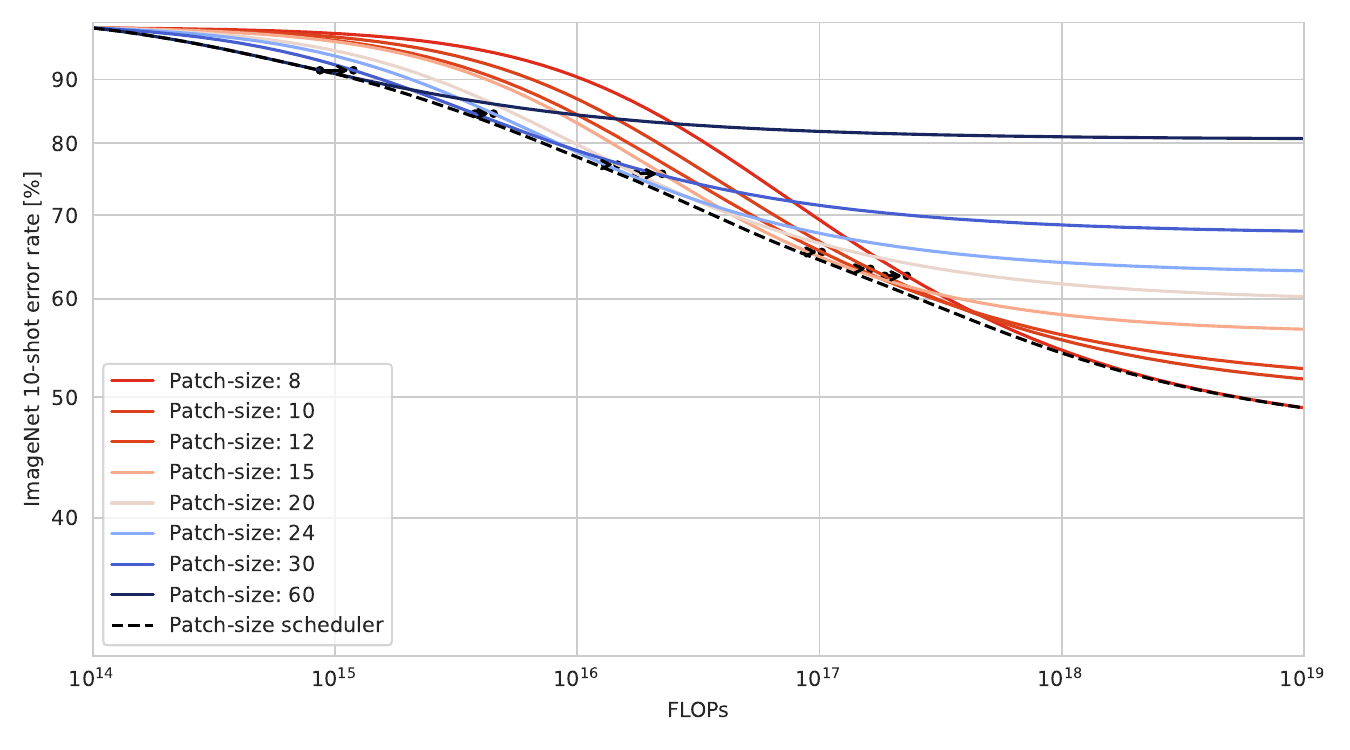} &   \includegraphics[width=0.48\linewidth]{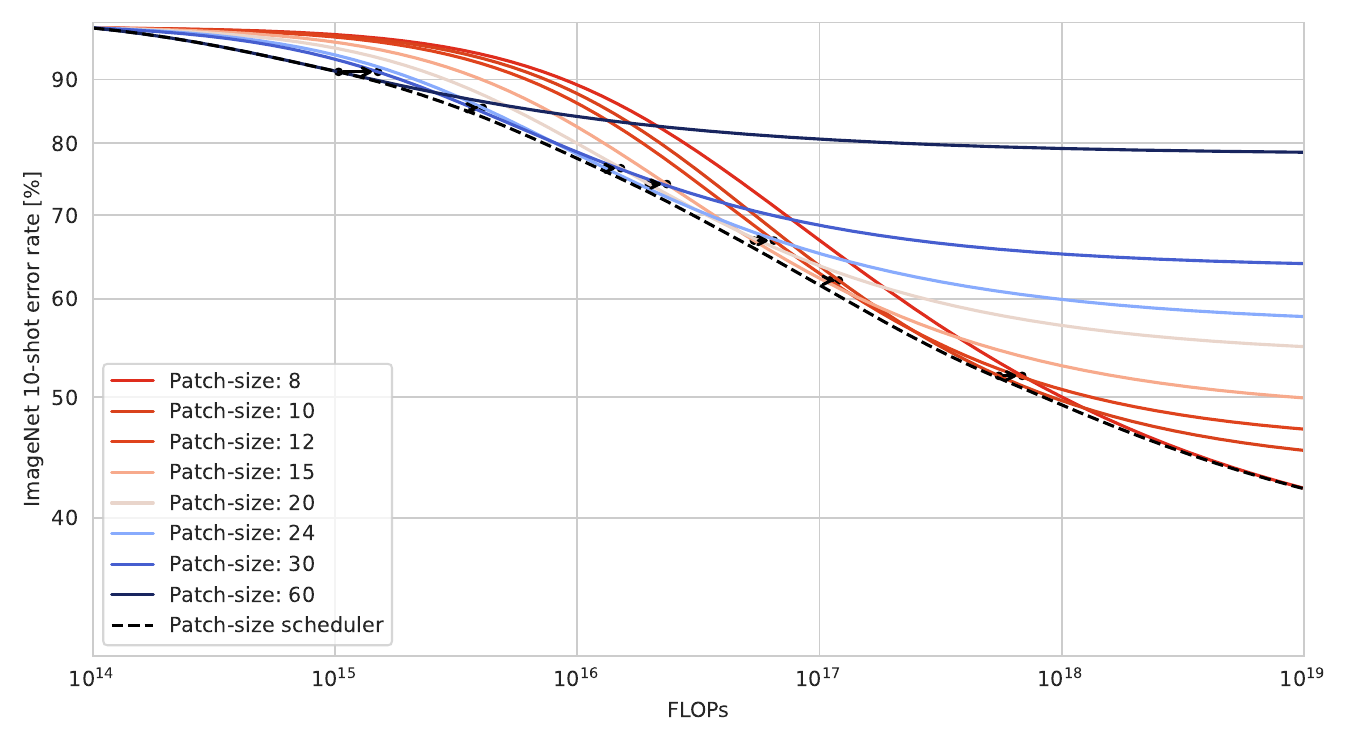} \\
(a) Model \textit{V}$256$\textit{-}$6$. & (b) Model \textit{V}$192$\textit{-}$12$. \\[6pt]
  \includegraphics[width=0.48\linewidth]{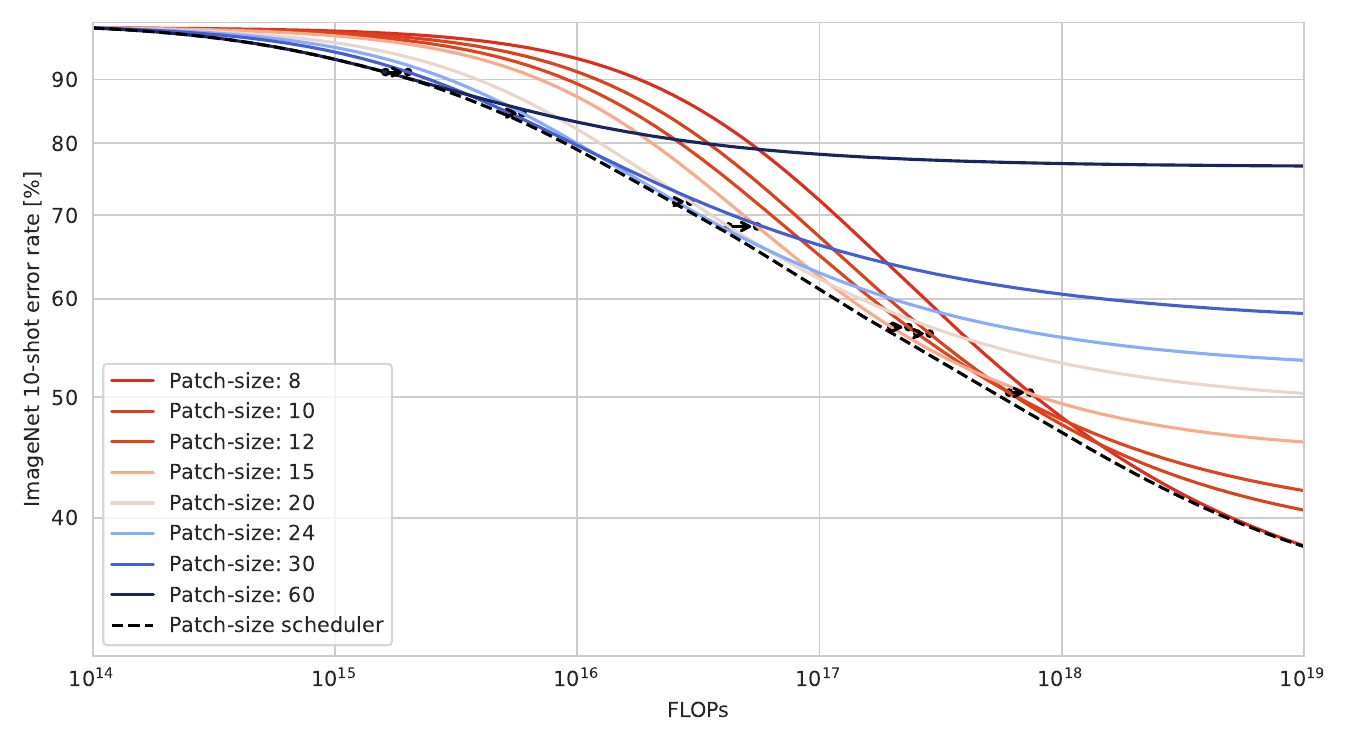} &   \includegraphics[width=0.48\linewidth]{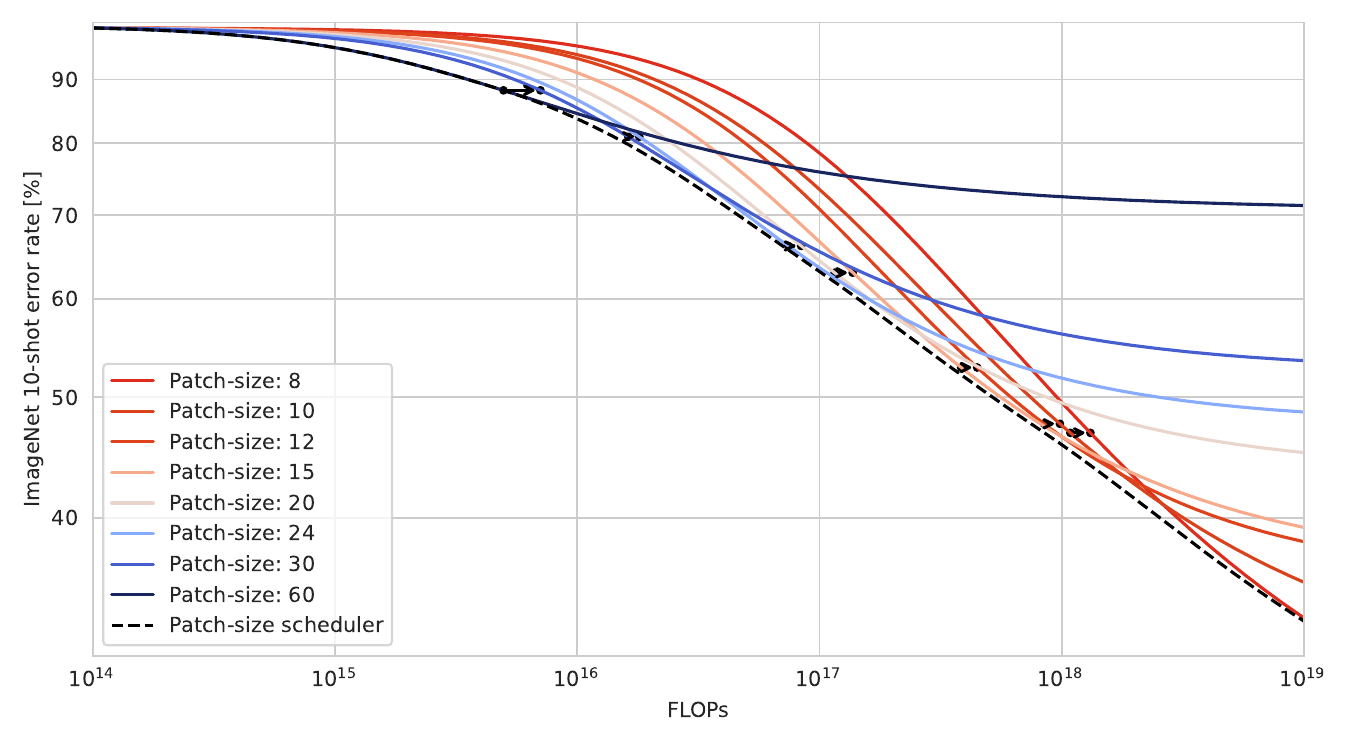} \\
(c) Model \textit{V}$256$\textit{-}$12$. & (d) Model \textit{V}$384$\textit{-}$12$. \\[6pt]
  \includegraphics[width=0.48\linewidth]{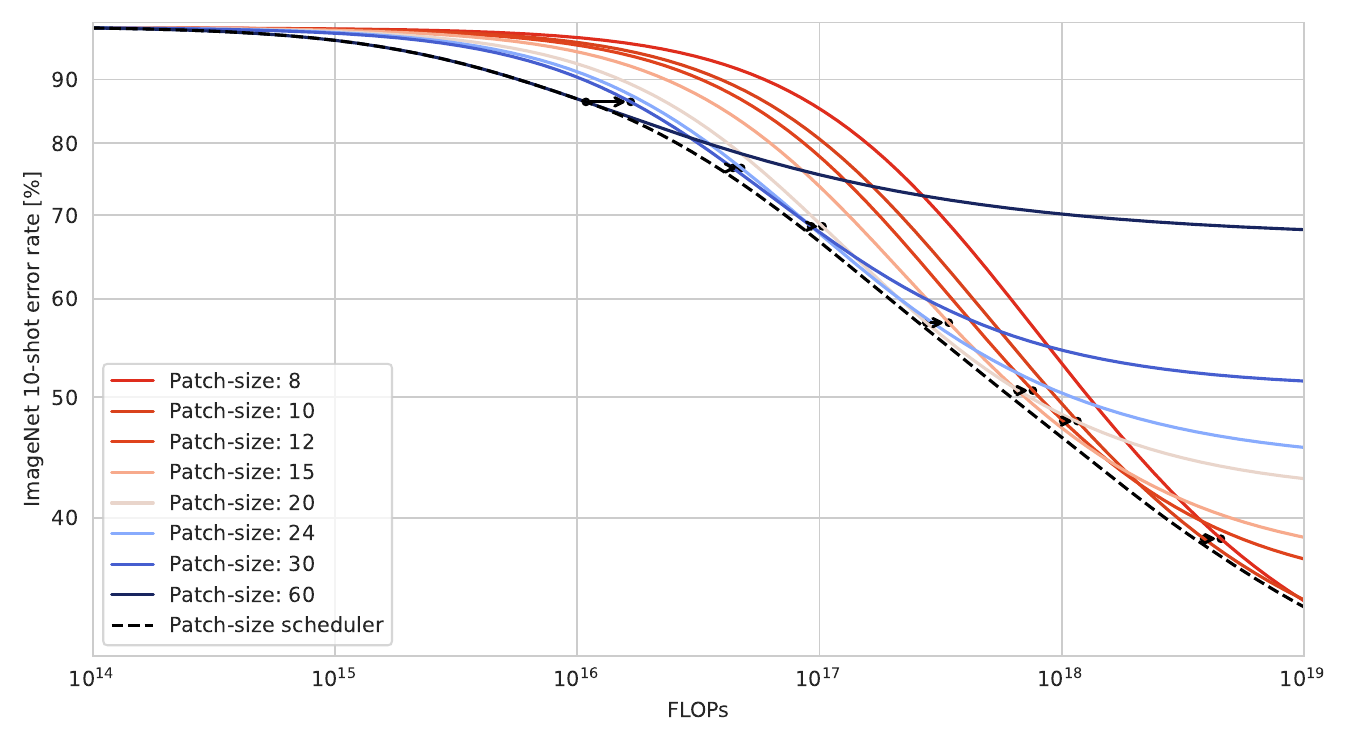} &   \includegraphics[width=0.48\linewidth]{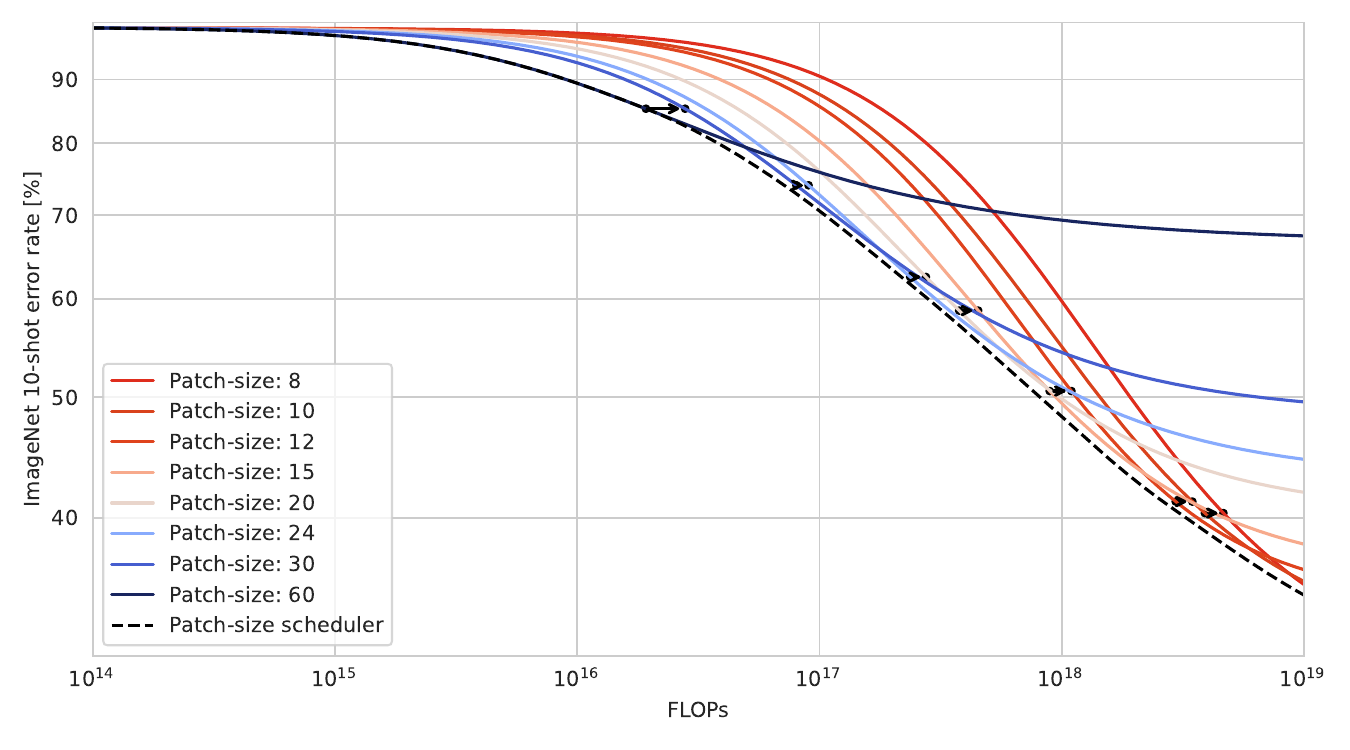} \\
(e) Model \textit{V}$512$\textit{-}$12$. & (f) Model \textit{V}$640$\textit{-}$12$. \\[6pt]
\end{tabular}
\caption{Fitted scaling laws and the predicted transition points that lead to the steepest descent.}
\label{fig:patch_size_scheduler_theory}
\end{figure*}

\begin{figure*}[!h]
\begin{center}
\begin{tabular}{cc}
  \includegraphics[width=1\linewidth]{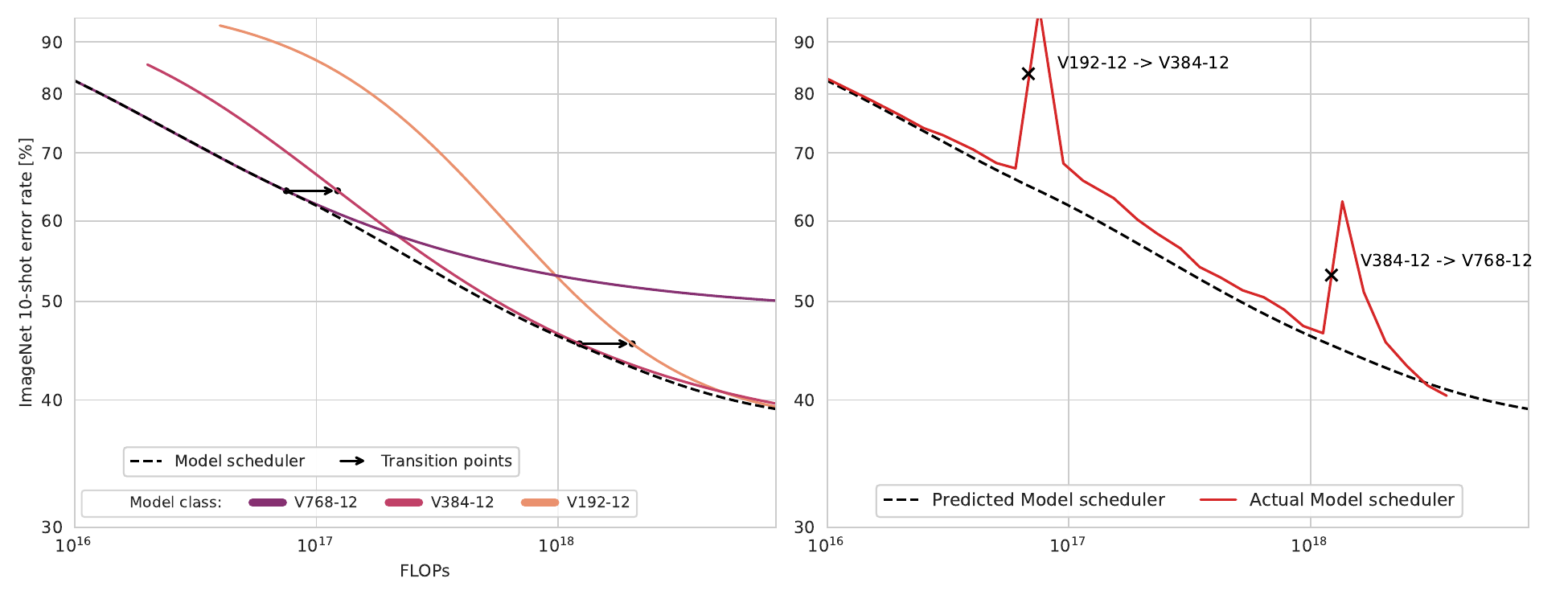} \\ 
  (a) Patch size $15$. \\[6pt]
  \includegraphics[width=1\linewidth]{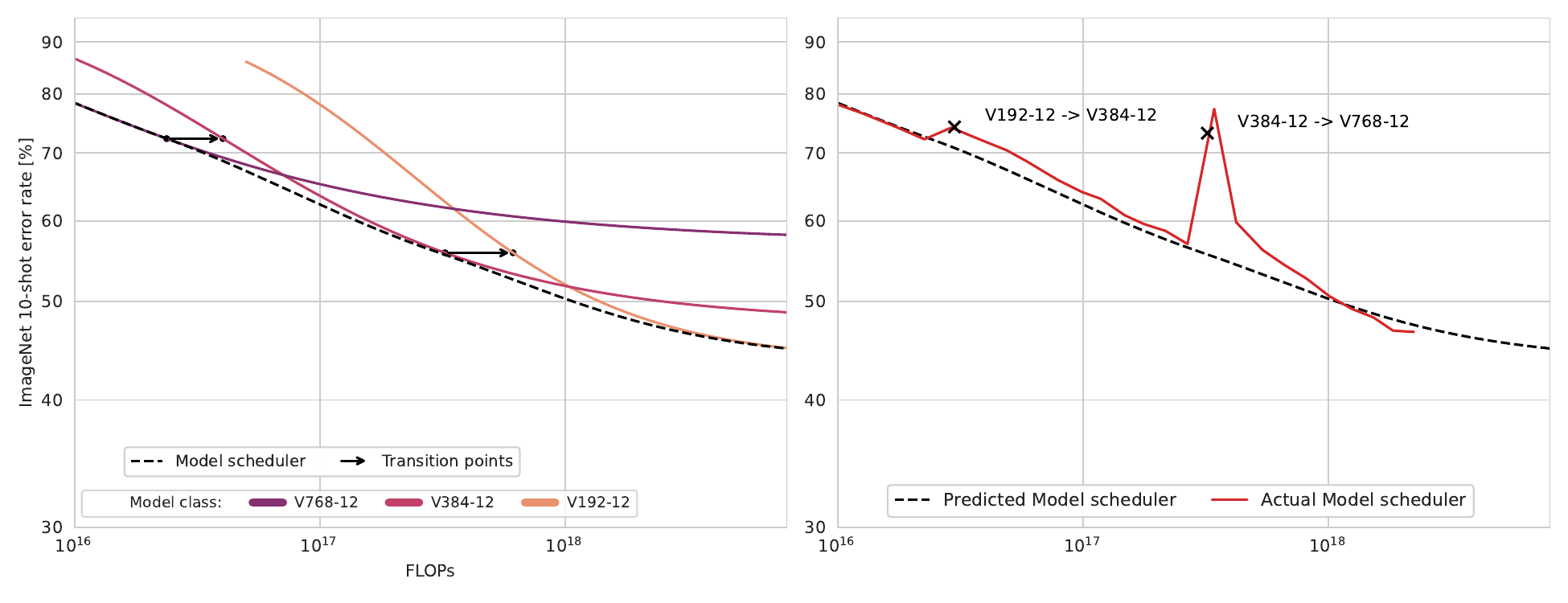} \\
  (a) Patch size $24$. \\[6pt]
  \includegraphics[width=1\linewidth]{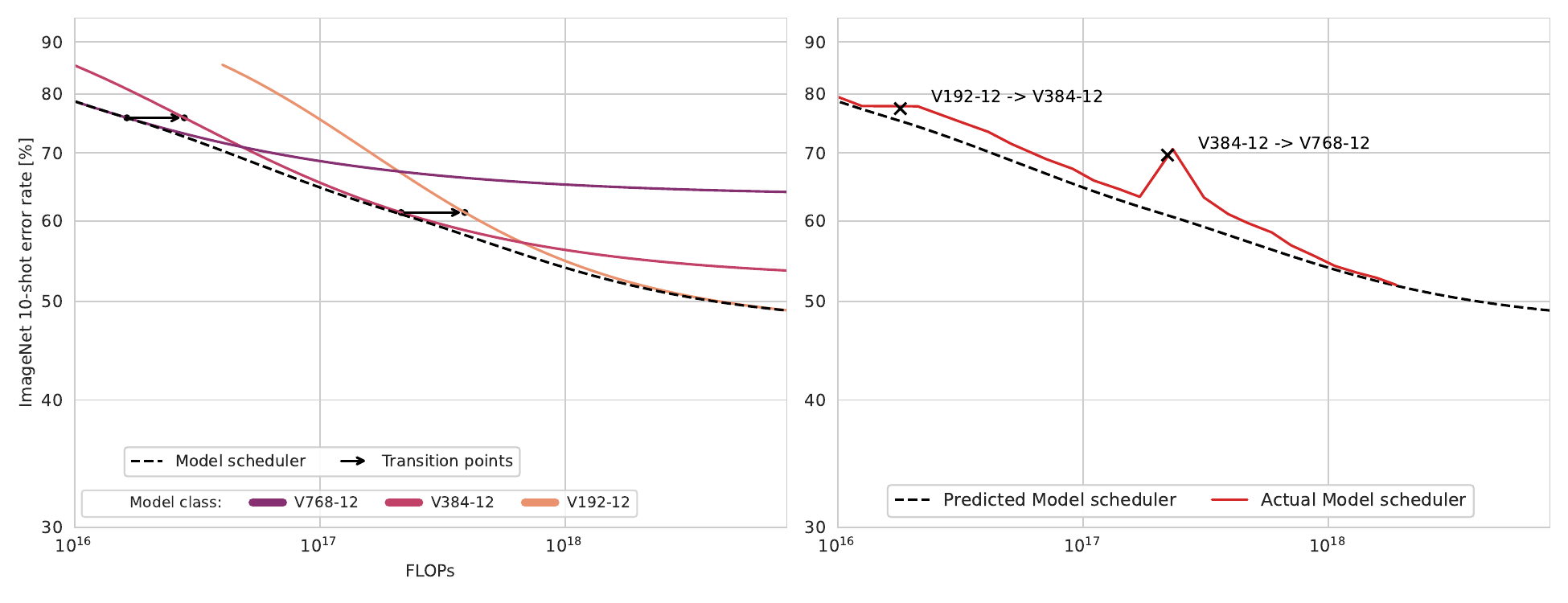} \\
  (c) Patch size $30$. \\[6pt]
\end{tabular}
\end{center}
\caption{Width scheduler for models trained with different patch sizes. We expand the model width twice, as done in Sec.~\ref{sec:adapting-width}. The transition points of the expansion are found through our maximum descent rule.}
\label{fig:additional_width_expansions}
\end{figure*}

\begin{figure*}
    \centering
    \includegraphics[width=0.8\linewidth]{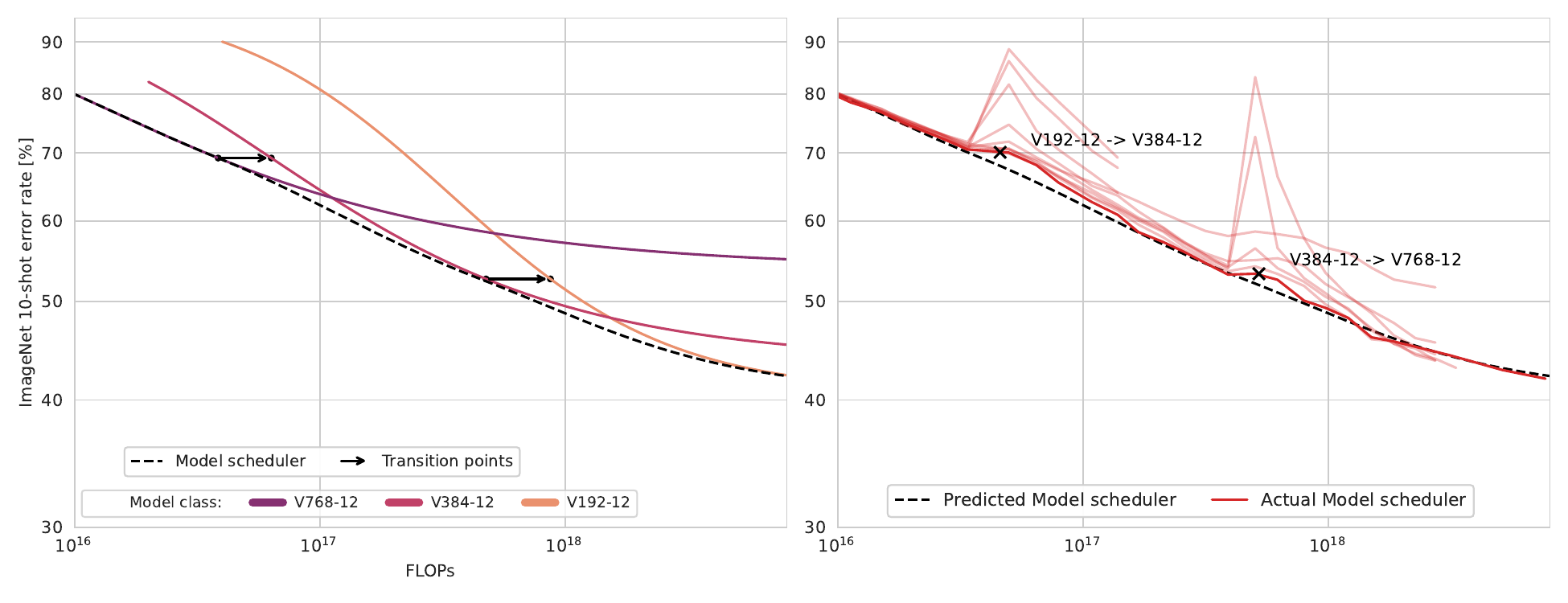}
    \caption{Different initialization schemes when expanding the width of the model. In practice, we set the variance of the new weights to be $\gamma \sigma^2$, where $\sigma^2$ is calculated from the pre-expanded weights $\mW$, for different values of $\gamma \in \{0.25, 0.5, 0.75, 1, 1.25, 1.5\}$.}
    \label{fig:dif-inits}
\end{figure*}

\paragraph{Patch size scheduler.} We present additional experiments on patch size schedulers in Fig.~\ref{ref:patch_size_scheduler_additional}. For \flexivit~-- similar to the original paper -- we sample uniformly at every step a patch size from the set $\{8, 10, 12, 15, 20, 24\}$. We did not use smaller patch sizes due to computational constraints. Note that our patch size scheduler leads to significantly faster convergence across the model classes we are analyzing. We also present in Fig.~\ref{fig:patch_size_scheduler_theory}, the fitted scaling curves and the points where changing the patch size leads to the steepest descent for different scaling laws.

\paragraph{Model width scheduler.} Supplementary to the results in Sec.~\ref{sec:adapting-width}, we provide additional examples of width examples in Fig.~\ref{fig:additional_width_expansions}. Note that we do not touch on the (1) \textit{where} to add the new weights and (2) \textit{how} to initialize these new weights. Our approach simly defines a strategy on the \textit{when} to expand the model and can be used in conjunction with any related works that provide answers to the previous (1) and (2) questions.

Regarding (1), we focus on models with constant depth (remember we are using the established \textit{Ti}, \textit{S}, and \textit{B} Vision Transformer sizes). Therefore, we do not add new layers but merely expand the weight matrices to the new embedding dimension. Our method is agnostic to \textit{where} these weights are added, just on the final form of the scaling law. Note that there exist more optimal ways to expand the different components of a ViT model~\citep{alabdulmohsin2023getting}.

Regarding (2), there are numerous works on \textit{how} to initialize the weights under a function preservation criterion. In our case, we found that zero-initializing weights, as commonly proposed, is significantly suboptimal. In practice, we expand the weights matrices by initializing the new entries in the weight matrices randomly based on the norm of the weights of the already learned weights. In more detail, linear layers are expanded as:
\begin{equation*}
\mW^\prime = \begin{pmatrix}
\mW & \mW_1 \\
\mW_2 & \mW_3,
\end{pmatrix}
\end{equation*}
where $\mW_1, \mW_2, \mW_3 \sim\mathcal{N}(\bm{0},\,\sigma^2 \bm{I})$, and $\sigma^2$ is calculated from $\mW$. This ensures better signal propagation in the network~\citep{he2015delving, noci2022signal}. The effect of this initialization can be important, but not detrimental, as illustrated in Fig.~\ref{fig:dif-inits}. When expanding the self-attention layers, we simply concatenate new heads, i.e. leave the heads that correspond to the previous embedding dimension unchanged. Again we stress that our method does not attempt to answer the question on \textit{how} to initialize, and any method established in the literature can be used for this purpose.

\begin{figure*}[h]
   \centering
   \includegraphics[width=1.0\linewidth]{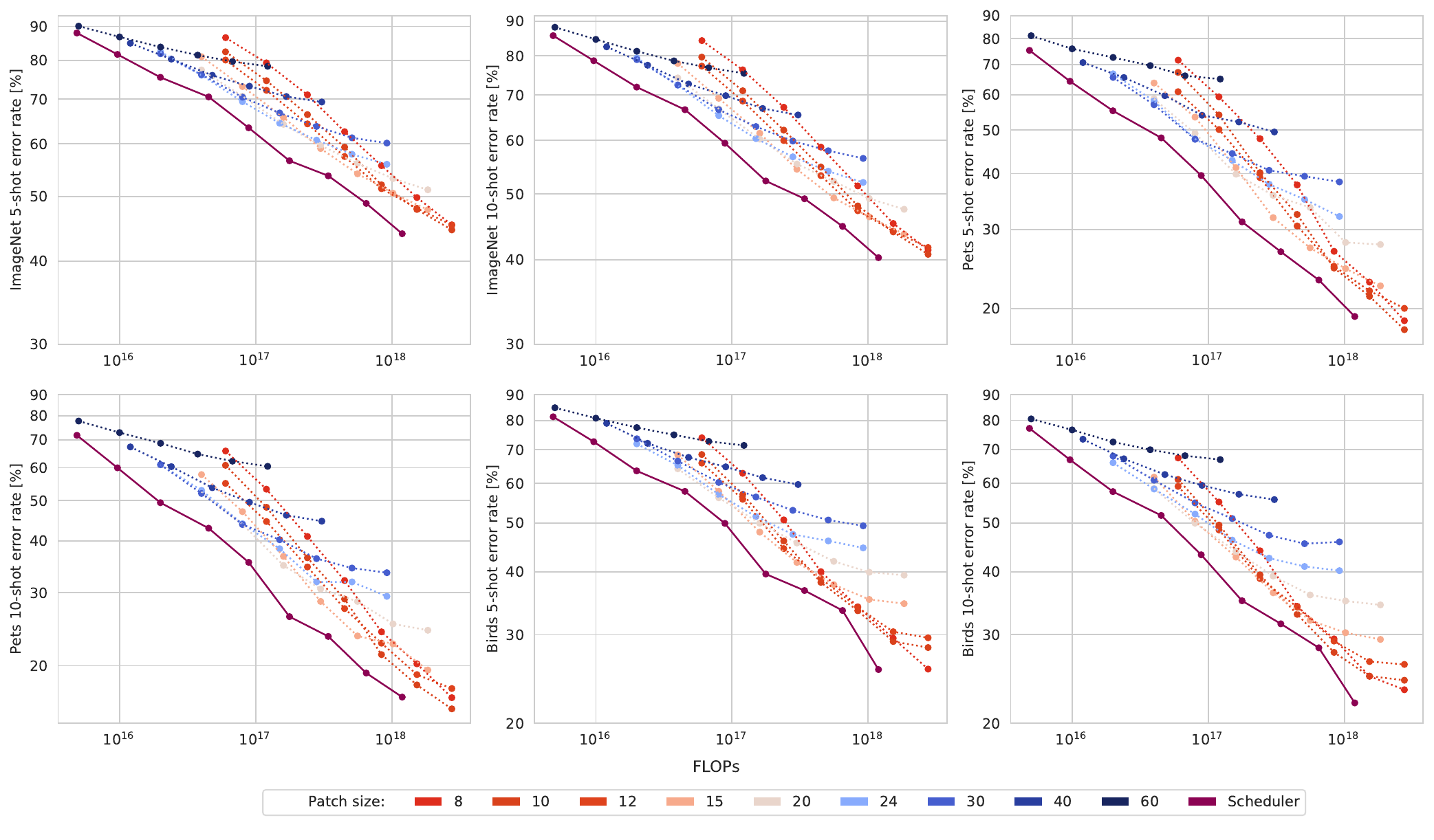}
   \caption{We include more downstream performance metrics for the \textit{V}$384$\textit{-}$12$ model.}
   \label{fig:more_downstream}
\end{figure*}

We also include additional commonly used downstream performance metrics. In particular, we report 5/10-shot results on \textit{ImageNet/Pets/Birds} as done in~\citet{zhai2022scaling}. Results can be seen in Fig.~\ref{fig:more_downstream}.

\section{Adapting Multiple Model Shape Parameters Concurrently}
\label{app:model_patch_together}
We first present more results on which model configuration (number of parameters or patch size) leads to the most efficient training for different levels of performance in Fig.~\ref{fig:varying_ps_ms} and~\ref{fig:imagenet_10shot_grad}.

Motivated by these insights, we ask the question: \textit{Can we change both the model size and patch size during training, leading to even greater training compute savings?}

\begin{figure}[h]
   \centering
   \includegraphics[width=1.0\linewidth]{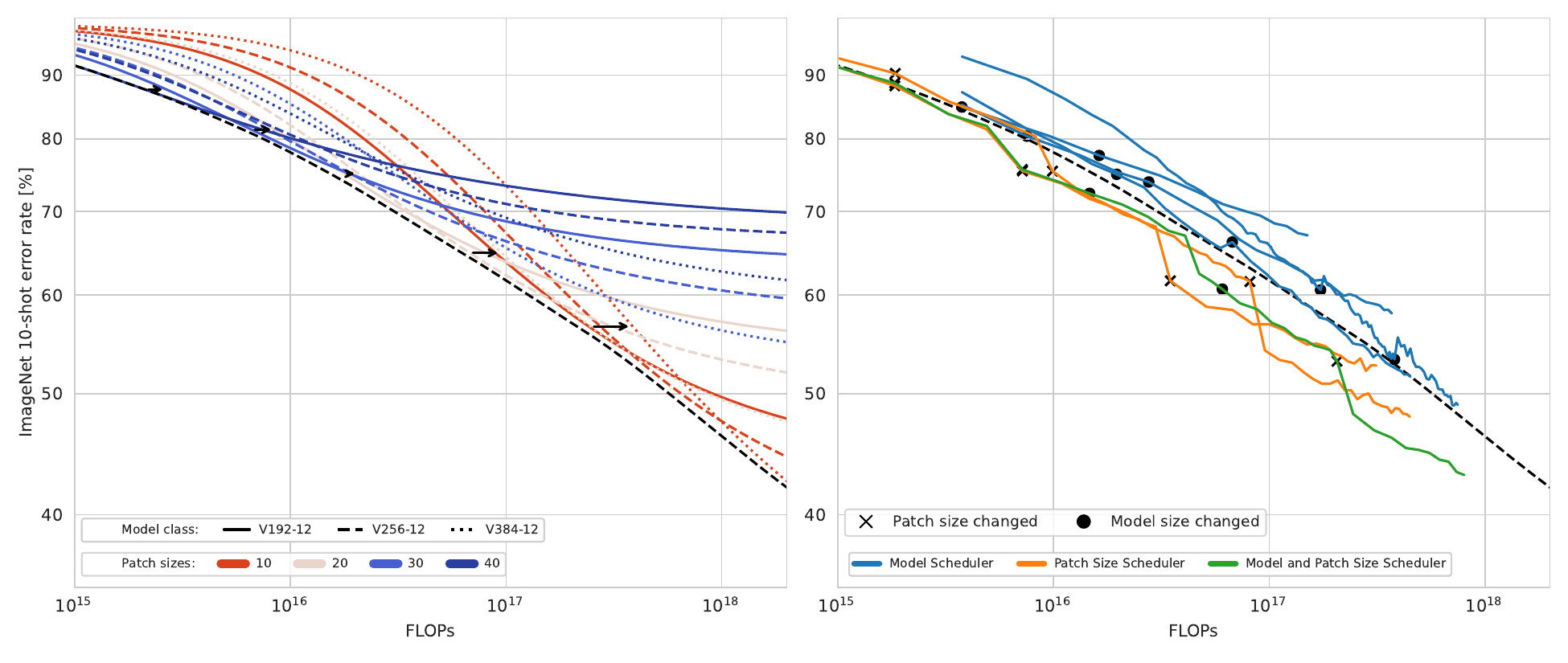}
   \caption{Changing both model width and patch size during training, further accelerates training.}
   \label{fig:ms_ps_scheduler}
\end{figure}

\begin{figure*}[t]
   \centering
   \includegraphics[width=1.0\linewidth]{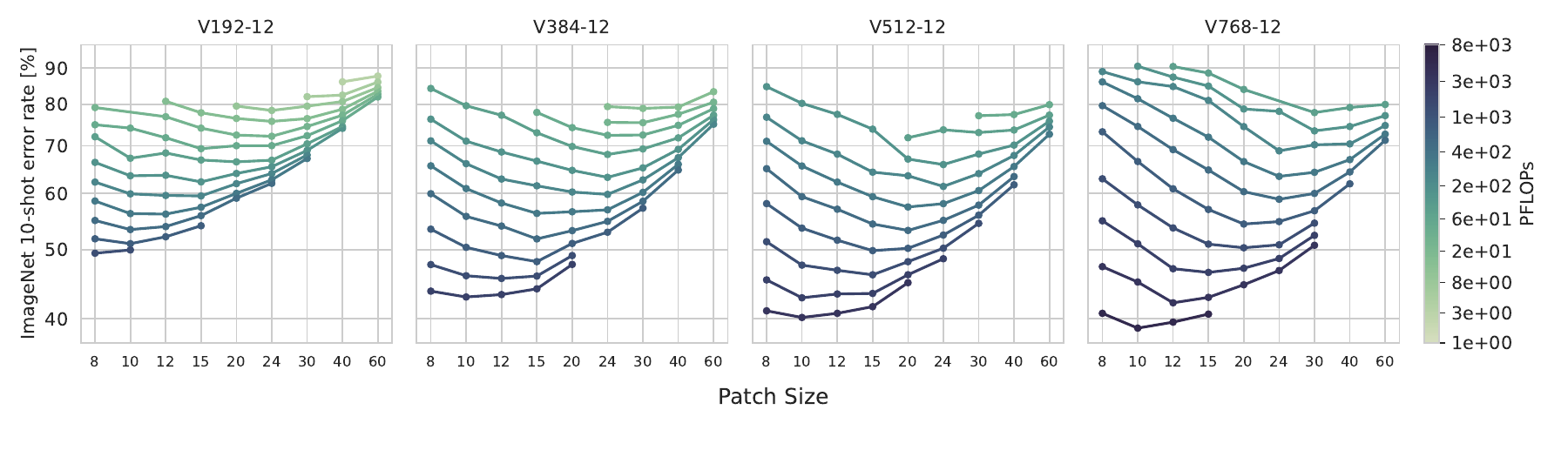}
   \caption{IsoFLOPs curves for different size ViTs trained with different constant patch sizes. Note how larger patch sizes are favored for smaller total FLOPs, while smaller patch sizes become more efficient as total FLOPs increase. Larger model sizes also become more favorable as total FLOPs increase.}
   \label{fig:varying_ps_ms}
\end{figure*}

\begin{figure*}[t]
   \centering
   \includegraphics[width=1.0\linewidth]{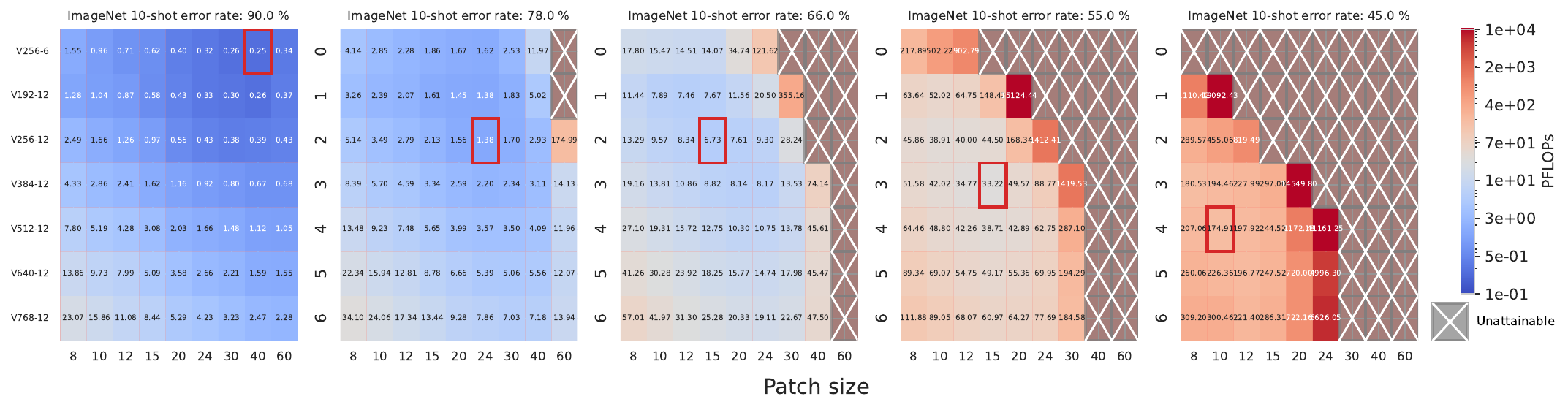}
   \caption{Values for $-\frac{\partial g_P(E)}{\partial E}$ in Eq.~\ref{eq:grad_equation}. Values indicate how many FLOPs are required for a proportionate increase in performance (i.e. drop in the error rate).}
   \label{fig:imagenet_10shot_grad}
\end{figure*}

We present preliminary experiments here, and more specifically in Fig.~\ref{fig:ms_ps_scheduler}. We compare results when changing only the model width, only the patch size, or both the model width and patch size simultaneously. In every case, we find the transition points, when the model shape should be adapted, using our proposed methodology. Changing both patch size and model width leads to the most significant improvements. For simplicity and clarity, we here consider model sizes in the set \{\textit{V}$192$\textit{-}$12$, \textit{V}$256$\textit{-}$12$, \textit{V}$384$\textit{-}$12$\} and patch sizes in the set \{10, 20, 30, 40\}.

We note that our method does not take into account momentary performance boost, when reducing the patch size and momentary performance deterioration when changing the model size, due to reasons highlighted in the main text. This justifies why changing only patch size can be better in some cases for the short term. As more compute is invested into the new model shape, these changes are counteracted.

\section{Environmental Impact}
\label{app:enviromental_impact}

To estimate the carbon footprint of our training, we follow the recipe detailed in \citet{touvron2023llama}. Specifically, we approximate the Watt-hours (Wh) used as
$$\text{Wh} = \text{GPU-hours} \times \text{GPU-power-consumption} \times \text{PUE}$$
where \text{PUE} refers to Power Usage Effectiveness. Following~\citet{touvron2023llama} we set this quantity to $1.1$. In order to enable comparisons across different works, we use the national US average carbon intensity factor of $0.385\hspace{1mm} kg\hspace{1mm}CO_2eq/KWh$ and we thus estimate the amounts of carbon emissions as 
$$tCO_2eq = MWh \times 0.385.$$
We compare our adaptively trained model against standard training of the compute-optimal model, in this case, the ViT Base model with patch size 8. The model requires $\approx 120$ GPU-hours with an average consumption of $\approx 280 W$ with the default training. Our adaptive training requires roughly $40\%$ of GPU-hours, i.e. $\approx 48$ GPU-hours while enjoying the same average consumption $\approx 280 W$.  This leads to $\approx 0.036MWh$ for ViT-Base and $\approx 0.014MWh$ for our adaptive training. Thus, the default training of the ViT Base model causes carbon emissions of $0.014tCO_2eq$ and our training $0.006tCO_2eq$.

\section{Time Measurement}
\label{app:time_measurement}

\begin{figure}[h]
   \centering
   \includegraphics[width=0.7\linewidth]{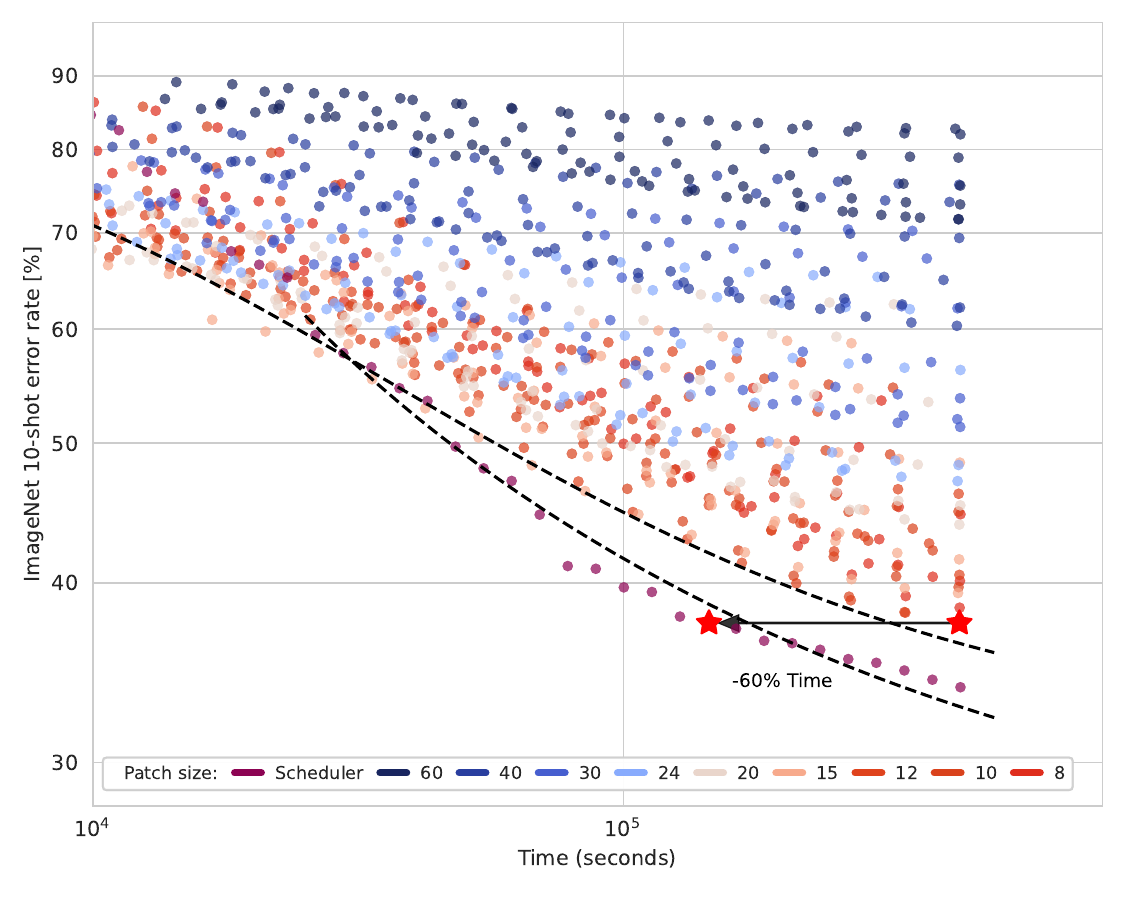}
   \caption{Same plot as Fig.~\ref{fig:optimality} but with time instead of FLOPs in the x-axis.}
   \label{fig:optimality_time}
\end{figure}

Although we focused on FLOPS, a similar hardware-aware analysis can take place, where the desired quantity to minimize is time instead of FLOPs. We note that time and FLOPs are usually highly correlated~\citep{alabdulmohsin2023getting}. This relationship also depends on the type of hardware and the mode it is operating in, i.e. whether we are memory-bound, whether data loading is the bottleneck etc. As an additional result, we replicate Fig.~\ref{fig:optimality} from the main text but with time on the x-axis. Results can be seen in Fig.~\ref{fig:optimality_time}.

\section{Discussion}

Although our approach is not directly comparable or inspired by them, we discuss some further interesting connections.

\paragraph{Neural Architecture Search.}~The discovery of optimal architectures has also been explored in the line of work of neural architecture search~\citep{elsken2019neural}. Neural architecture search explores a collection of techniques to automate the selection of an optimal architecture. We are interested in more efficient training for a fixed architecture, the Transformer, that has established itself across different modalities.

\paragraph{Curriculum Learning.}~Curriculum learning argues that the order in which samples are presented plays a crucial role in the learning efficiency of a model~\citep{soviany2022curriculum}. The role of training data undoubtedly played a crucial role in the convergence speed~\citep{sorscher2022beyond}. Recently, other techniques for data selection have been proposed to accelerate large-scale pre-training~\citep{evans2023bad}. Our technique does not filter or select data, just chooses to invest different amounts of compute to different data, based on the current stage of training. Population based training is also a related area of work~\citep{jaderberg2017population}.

\end{document}